\pdfoutput=1
\documentclass[10pt, logo, twocolumn, copyright]{nvidiatechreport}

\usepackage{hyperref}
\usepackage{cleveref}
\crefname{figure}{Figure}{Figures}\Crefname{figure}{Figure}{Figures}
\crefname{table}{Table}{Tables}\Crefname{table}{Table}{Tables}
\crefname{section}{Section}{Sections}\Crefname{section}{Section}{Sections}
\crefname{appendix}{Appendix}{Appendices}\Crefname{appendix}{Appendix}{Appendices}
\crefname{equation}{Eq.}{Eqs.}\Crefname{equation}{Equation}{Equations}
\crefname{algorithm}{Algorithm}{Algorithms}\Crefname{algorithm}{Algorithm}{Algorithms}
\crefname{proposition}{Proposition}{Propositions}\Crefname{proposition}{Proposition}{Propositions}
\usepackage{url}
\usepackage{xurl}
\usepackage[dvipsnames]{xcolor}
\usepackage[table]{xcolor}
\usepackage{colortbl}
\usepackage{nicefrac}
\usepackage{microtype}
\usepackage{graphicx}
\usepackage{subcaption}
\usepackage{caption}
\usepackage{float}
\usepackage{placeins}
\usepackage{stfloats}
\usepackage{wrapfig}
\usepackage{amsmath}
\usepackage{amssymb}
\usepackage{multirow}
\usepackage{booktabs}
\usepackage{makecell}
\usepackage{arydshln}
\usepackage{adjustbox}
\usepackage{enumitem}
\usepackage{xspace}
\usepackage{listings}
\usepackage{mdframed}
\usepackage{tcolorbox}
\usepackage{color,soul}
\usepackage[numbers]{natbib}
\usepackage{amsthm}
\usepackage{algorithm}
\usepackage{algpseudocode}
\usepackage{array,booktabs,colortbl,xcolor}
\newtheorem{proposition}{Proposition}

\tcbuselibrary{listings,breakable}

\definecolor{TitleCardGreen}{HTML}{EFF7DF}
\definecolor{NvidiaGreen}{RGB}{118, 185, 0}
\sethlcolor{red!15}

\expandafter\let\csname endtabular*\endcsname\endtabular

\newcommand{\shortname}{LoRN\xspace}
\newcommand{\longname}{Low-Rank Noise Steering\xspace}

\newif\ifshowannotations
\showannotationstrue

\title{\shortname: Low-Rank Noise Steering for Generative Robot Policies}

\author[1*]{Wei Liu}
\author[1*]{Abrar Anwar}
\author[1]{Huihua Zhao}
\author[1]{Soha Pouya}
\author[1]{Yan Chang}
\affil[1]{NVIDIA}
\affil[*]{Equal contribution}
 
\newcommand{\projectpage}{https://nvidia-isaac.github.io/isaac-rlpt/lorn/}

\begin{abstract}
Noise steering improves pretrained generative robot policies by learning to select their initial noise without updating model weights.
We find that actions in the evaluated world-action models are more sensitive to video noise than action noise, making video noise an expressive input for steering.

However, the noise-steering methods we evaluate, in their published forms, struggle to scale to the high-dimensional noise inputs of large policy classes such as world-action models.
We introduce Low-Rank Noise Steering (\shortname), a reinforcement learning method that scales noise steering to world-action models.
A small steering actor selects a direction through a fixed low-rank random projection, while our norm-controlled construction keeps the resulting noise near the typical scale of a Gaussian sample.
This separates the actor's control dimension from the full noise dimension, enabling steering of both video and action inputs.
Across three world-action models and ten model--task pairs, video steering improves average success by 26.2 percentage points over the base policy, compared with 7.1 points for DSRL action steering.
For smaller action policies, the same construction achieves comparable final success while generally improving early-training success. Steering actors trained in simulation also transfer to three physical bimanual manipulation tasks across two base policies, without additional real-world training.
\end{abstract}

\newcommand{\teaserfigure}{%
    \begin{center}
        \includegraphics[width=1\textwidth]{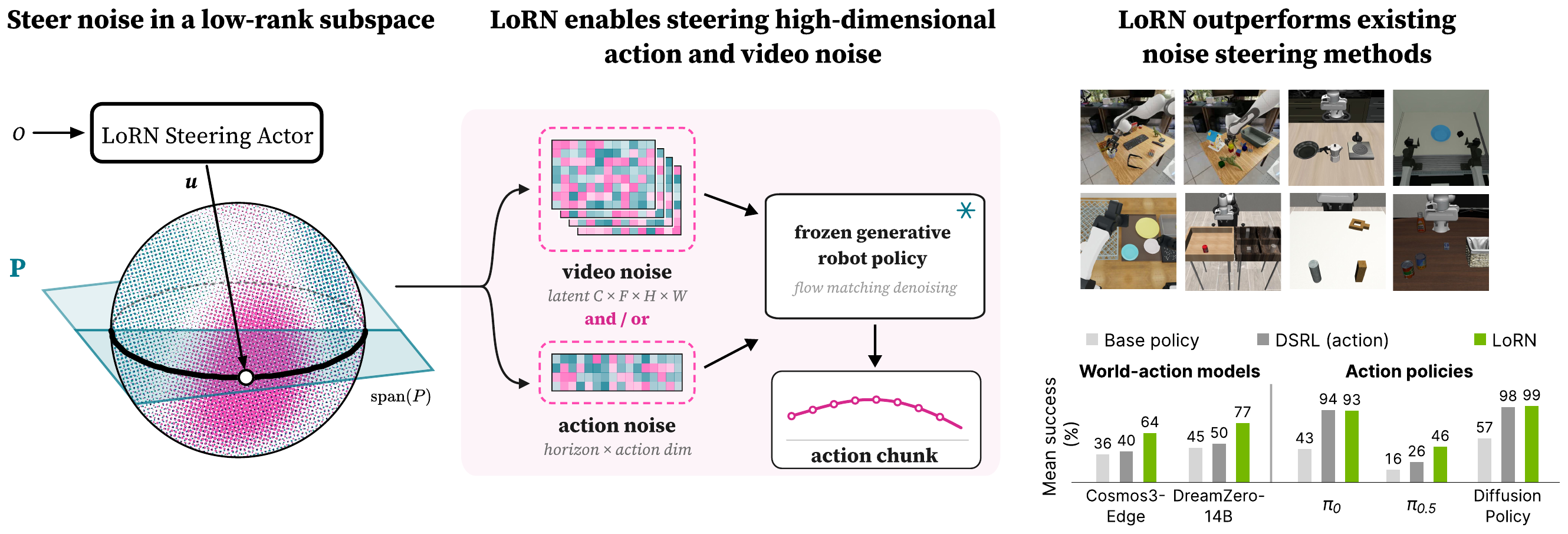}
        \captionof{figure}{
\textbf{LoRN enables low-rank noise steering of large generative robot policies.}
Existing noise steering methods improve frozen policies by steering the initial noise used to generate an action chunk.
However, extending noise steering to the much higher-dimensional outputs of vision-language-action models and world-action models is challenging.
\textbf{Left:} LoRN learns a small set of coefficients that selects a noise direction within a fixed low-rank subspace, keeping the actor's output dimension small regardless of the full noise dimension.
\textbf{Middle:} This formulation allows LoRN to steer either action noise or video noise in world-action models. 
We find that steering the video noise is surprisingly expressive for denoising actions, thus LoRN enables a new way to steer a world-action model's actions.
\textbf{Right:} Mean final success per policy family, averaged over its tasks, for the base policy, DSRL action-noise steering, and LoRN.
Video-noise steering produces large gains on the evaluated world-action models; on action policies, LoRN achieves comparable final success to DSRL while generally improving early-training success (\cref{tab:action}).
The DSRL bars use the same variants as \cref{fig:wam-success}, with its full-dimensional form on Cosmos3-Edge.
}
        \label{fig:teaser}
    \end{center}
}

\let\abstractwithoutteaser\abscontent

\fancypagestyle{titlecardstyle}{%
    \fancyhf{}%
    \fancyfoot[L]{\copyrightext}%
    \fancyfoot[R]{\footerfont\thepage}%
}

\makeatletter
\newcommand{\titlecardcontent}{%
    \begin{tcolorbox}[
        colback=TitleCardGreen,
        colframe=TitleCardGreen,
        boxrule=0pt,
        arc=4mm,
        outer arc=4mm,
        boxsep=0pt,
        left=7mm,
        right=7mm,
        top=6mm,
        bottom=2mm
    ]
        {\raggedright\color{black}\sffamily\bfseries\fontsize{21}{24}\selectfont\@title\par}
        \vskip15pt
        {\raggedright\@author\par}
        \vskip13pt
        \abstractwithoutteaser
        \vskip6pt
        {\raggedright\small\sffamily\textbf{Project page:} \url{\projectpage}\par}
        \vskip4pt
        \hfill\includegraphics[width=78pt,trim=303bp 303bp 303bp 304bp,clip]{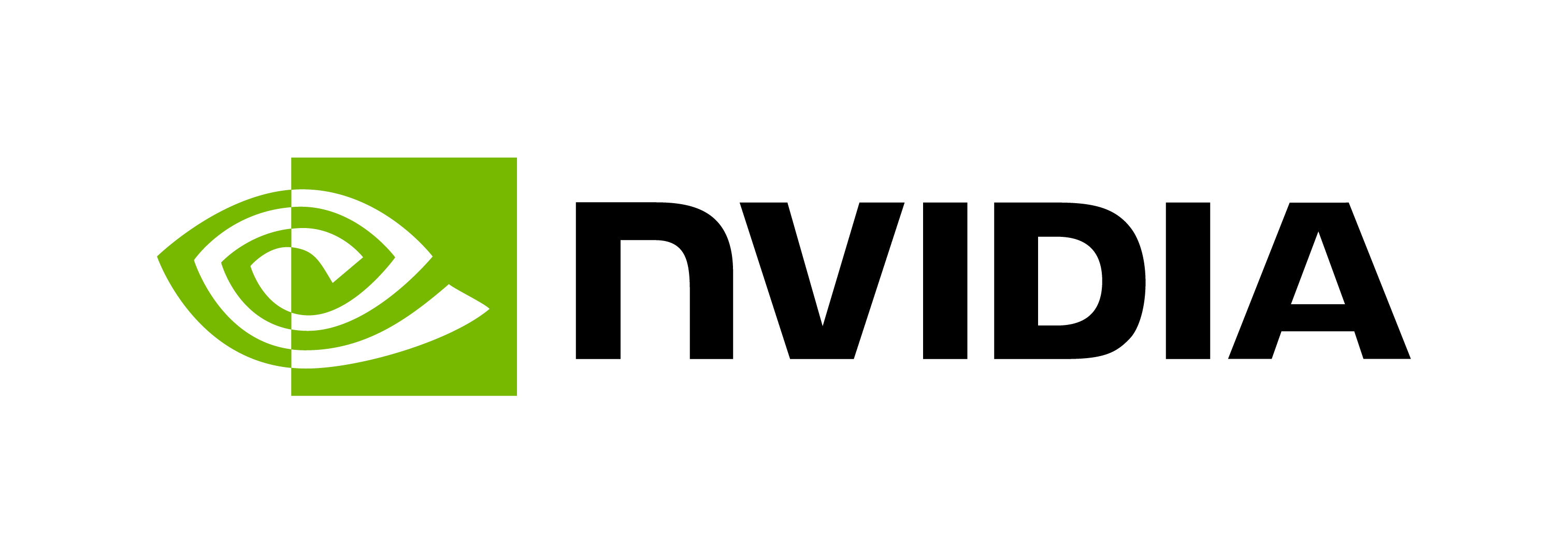}
    \end{tcolorbox}
    \vskip2pt
    \teaserfigure
}
\renewcommand{\maketitle}{%
    \begingroup
    \addtolength{\topmargin}{-30pt}%
    \addtolength{\textheight}{30pt}%
    \onecolumn
    \thispagestyle{titlecardstyle}%
    \titlecardcontent
    \clearpage
    \endgroup
    \twocolumn
}
\makeatother

\begin{document}

\maketitle

\section{Introduction}
\label{sec:intro}

Large robot foundation models, such as vision-language-action models (VLAs) and world-action models (WAMs), can perform a wide range of behaviors.
For these models to reliably succeed on a particular task, they must learn to improve via interaction with their environment.
Reinforcement learning (RL) offers a way to improve their task performance through interaction, but updating large policies can be computationally expensive.
Noise steering has emerged as an approach to guide the denoising process of diffusion and flow-matching policies by replacing the initial noise samples. 
However, these methods struggle to scale to high-dimensional noise inputs that are common in large policy classes such as VLAs and WAMs. 
This work develops a noise steering method that addresses this challenge, focusing primarily on WAMs.

World-action models couple visual dynamics prediction with action generation.
Their sampling inputs include initial source distribution noise associated with both video and actions, and changing the video noise can also change the actions the model produces. 
Existing action-noise steering methods do not exploit this input, and in the implementations and settings we evaluate they give no consistent improvement when applied to the substantially larger video-noise space.
In this work, we find that WAMs' decoded actions are \textit{more sensitive} to video noise than action noise, implying video noise may serve as a strong steering modality.

We introduce \longname (\shortname), a method to efficiently train noise steering policies for world-action models.
Recent work has found that selecting the initial noise samples of diffusion or flow matching heads can effectively steer robot policies in a sample efficient manner without requiring any updates to the full model~\citep{dsrl}.
We find that extending noise steering to large robot foundation models often fails, primarily because their noise inputs are high-dimensional and their sensitivity to it differs sharply between the video and action inputs.

\shortname overcomes these challenges by steering a low-rank projection of the noise distribution. 
A steering actor outputs a low-dimensional vector that a fixed random projection lifts to the full noise dimension, which is then rescaled to resemble a Gaussian sample's norm.
The frozen policy denoises this noise, and the actor is trained with reinforcement learning.
This formulation scales to high-dimensional action spaces and, most surprisingly, enables the steerability of the \textit{video} noise to improve denoised actions.

Two design choices underlie our approach.
First, the actor controls only a small, low-rank subspace of the noise regardless of the input dimension. 
We show that such a low-rank subspace contains directions that improve on the base policy with high probability.
Second, mixing the actor's direction with a fresh Gaussian sample preserves the expected squared norm of the noise, so the steered noise stays near the typical norm that diffusion and flow-matching policies are trained to denoise.
Because neither choice depends on the structure of the noise input, \shortname can be applied to either the video or the action noise of a WAM.
This flexibility matters in practice: we find that a WAM's actions are far more sensitive to one noise input than the other, and a probe that needs no reward identifies that input before training.
On the three models we evaluate, it is the video noise, which existing methods do not steer.

We evaluate \shortname on three world-action models, on action-noise steering with $\pi_0$, $\pi_{0.5}$, and Diffusion Policy, and on a physical bimanual robot station. 
In aggregate across three WAMs and ten model--task pairs, DSRL on the action noise improves the base policies by 7.1 percentage points, while \shortname steers the more expressive video noise and improves them by 26.2 percentage points.
We also evaluate \shortname on twelve action-policy simulation tasks, where it generally achieves higher early-training success against DSRL~\cite{dsrl} and LP-DS~\cite{lpds}.
Finally, we demonstrate that \shortname trained in simulation transfers to three physical bimanual manipulation tasks on two base policies, the released GR00T N1.7 and a $\pi_{0.5}$ fine-tuned in simulation, without additional training on hardware.

Our contributions are fourfold:
\begin{enumerate}
\item We introduce \shortname, low-rank noise steering for robot foundation models: a fixed random subspace of the noise, mixed with a fresh Gaussian sample, independent of the actor's output, so that the expected squared noise norm is preserved.
\item We identify noise-norm preservation as a key requirement for reliable steering, and measure the band of norms that each frozen policy tolerates.
\item We prove that a random subspace of rank $r$ misses the high-advantage noise of a state with probability at most $(1-p)^r$, where $p$ is that noise's share of the sphere, and that a policy which selects an improving direction in that subspace whenever one exists improves on the base policy.
\item We provide a practical recipe for steering world-action models, with a reward-free probe that finds which noise input to steer, and evaluate it on three world-action models, twelve action-policy tasks, and a bimanual robot station with two base policies.
\end{enumerate}
\section{Related Work}
\label{sec:related_work}

\paragraph{Reinforcement learning for generative robot policies.}
Diffusion and flow-matching policies~\citep{diffusionpolicy,pi0,flowmatching} generate action chunks by iteratively denoising a Gaussian sample.
The same principle underlies generalist models~\citep{groot} and world-action models, which jointly denoise future video and actions~\citep{cosmospolicy,dreamzero}.
These policies are commonly improved by fine-tuning their weights with policy gradients~\citep{dppo,reinflow} or by learning residual corrections to their actions~\citep{policydecorator}.
For models with billions of parameters, however, online weight updates are computationally expensive and risk eroding pretrained capabilities.
In contrast, \shortname keeps the policy frozen and adapts only its initial noise.

\paragraph{Steering frozen generative policies.}
DSRL~\citep{dsrl} showed that a frozen diffusion or flow policy can be adapted by training an RL agent in its noise space.
Concurrent work~\citep{nsl} likewise trains a Gaussian actor over the action noise of a frozen flow-matching VLA.
Subsequent work regularizes the learned noise with a budget on its deviation from a Gaussian sample~\citep{lpds}, or restricts the actor to the noise directions to which the decoder is most sensitive~\citep{ni2026pss}.
Golden-ticket search~\citep{goldenticket} instead finds a single fixed noise that improves a policy across states.
Latent policy steering~\citep{lps} fixes the norm of the steered noise at its typical value and trains the actor offline with gradients through a one-step policy.
These methods were developed for policies with low-dimensional noise.
They either do not explicitly control the norm of the steered noise~\citep{dsrl,ni2026pss,nsl}, control it only on average~\citep{lpds}, or require a sensitivity probe whose cost grows linearly with the noise dimension~\citep{ni2026pss}.
Fixing the norm exactly~\citep{lps} still leaves the full noise dimension as the search space and requires a policy that is differentiable in one step.
In contrast, \shortname steers a random low-rank slice that requires no probing or gradients through the policy and preserves the expected squared norm of the noise by construction, which allows it to scale to the video noise of world-action models.

\paragraph{Noise geometry and low-rank adaptation.}
High-dimensional Gaussian noise concentrates near a fixed norm.
Image generative models exploit this property by interpolating along the corresponding sphere~\citep{white2016,ddim} and regularizing the norm during noise optimization~\citep{reno,samuel2023}, and their outputs degrade when sampling drifts from it~\citep{ning2024}.
Separately, LoRA~\citep{lora} and studies of intrinsic dimension~\citep{intrinsicdim} show that large networks can be adapted within low-rank or random subspaces of their parameters.
Outside robotics, active-subspace noise steering~\citep{asns} searches the initial noise of a frozen flow model along a few objective-sensitive directions estimated from objective values alone.
\shortname brings both ideas to policy adaptation: it applies a random low-rank subspace to the \textit{input} of a frozen model rather than its weights, and keeps the result concentrated near the typical Gaussian norm.
\section{Preliminaries and Problem Statement}
\label{sec:problem}

We consider adapting a pretrained robot policy to a target task through online interaction.
Let $F(o, z)$ denote a frozen diffusion or flow-matching policy that maps an observation $o$ and an initial noise $z \in \mathbb{R}^N$ to an action chunk $a$.
Typically, during training of a policy $F$, $z$ is always drawn from the Gaussian source distribution, $z \sim \mathcal{N}(0, I_N)$.

We model the target task as a Markov decision process with observations $o \in \mathcal{O}$, action chunks $a \in \mathcal{A}$, transition dynamics $P$, reward function $r$, and discount factor $\gamma$.
DSRL noise steering~\citep{dsrl} replaces the input Gaussian sample $z$ with the output of a learned actor $\pi_\theta(z \mid o)$ and executes $a = F(o, z)$.
Because $F$ is fixed, it can be absorbed into the environment, which produces an MDP whose action space is the noise space $\mathbb{R}^N$.
The objective is to find an actor that maximizes the expected discounted return $J(\theta) = \mathbb{E}\big[\sum_t \gamma^t r(o_t, F(o_t, z_t))\big]$, where $z_t \sim \pi_\theta(\cdot \mid o_t)$.
DSRL instantiates this formulation with SAC~\citep{sac}, whose bounded output replaces the noise; for action-chunk policies such as $\pi_0$, it tiles a low-dimensional output across the steps of the chunk.

A standard Gaussian sample $z \in \mathbb{R}^N$ has norm $\|z\|$ close to $\sqrt{N}$, and by the Gaussian annulus theorem~\citep{blum2020foundations} nearly all samples lie in a thin shell around the sphere of that radius, whose width relative to the radius shrinks as $N$ grows.
We express norms as multiples of the typical norm $\sqrt{N}$.
On this scale, the norms of Gaussian samples vary by approximately $1/\sqrt{2N}$: about 13\% for the 28-dimensional action noise of a small diffusion policy, but only 0.1\% for the 506{,}880-dimensional video noise of Cosmos3-Edge.
A large policy is therefore pretrained only on noise from a very narrow range of norms. \cref{sec:analysis} examines how it behaves outside this range.

\begin{figure}[t]
    \centering
    \includegraphics[width=\linewidth]{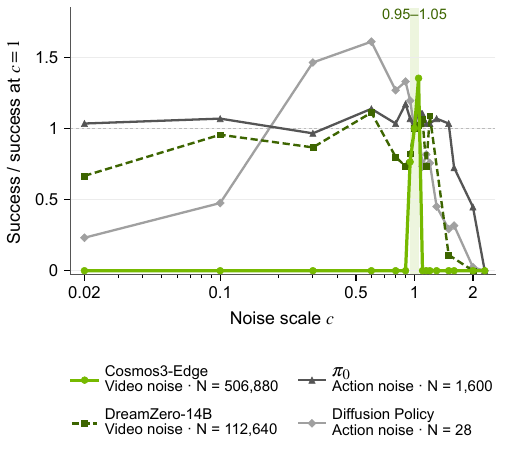}
    \caption{\textbf{Frozen policies tolerate only a band of noise norms.} Success when Gaussian samples are scaled by $c$, relative to $c = 1$. The shaded region marks the narrow band of Cosmos3-Edge.}
    \label{fig:bands}
\end{figure}

\section{Why Noise Steering Fails on Large Policies}
\label{sec:analysis}

Before introducing \shortname, we examine why noise steering, which is effective on small diffusion policies~\citep{dsrl}, fails on large robot foundation models.
We study four frozen policies: a Robomimic diffusion policy~\citep{diffusionpolicy,robomimic}, $\pi_0$~\citep{pi0}, and two world-action models, Cosmos3-Edge and DreamZero-14B~\citep{dreamzero}.
Their noise dimensions range from 28 for the Robomimic action noise to 506{,}880 for the Cosmos3-Edge video noise.
We identify two failure modes.

\paragraph{Failure mode 1: the search space grows with the noise dimension.}
The action noise of a generalist policy spans an entire chunk of padded action channels, and the video noise of a world-action model spans every latent pixel of every predicted frame.
A noise steering actor that sets each entry independently must search a space of dimension $N$, which reaches around $5 \times 10^5$ for the Cosmos video noise.
Steering such large policies therefore motivates the need for a search space whose dimension is independent of $N$.

\paragraph{Failure mode 2: the tolerated band of noise norms is policy-dependent and can be very narrow.}
We analyze how different input noise samples change policy performance by scaling Gaussian samples by a factor $c$ and measure each policy's success (\cref{fig:bands}). 
This scaling also scales the norm of the sample.
Every policy fails once $c$ moves far enough from $c=1$.
Cosmos3-Edge fails outside $0.95 \leq c \leq 1.05$, whereas DreamZero-14B holds its success from $c = 0.1$ to $1.2$ and fails only by $c = 1.5$.
The noise-steering methods we compare against leave the norm uncontrolled or constrain it only on average, and often fall outside this band. 
For example, we find that DSRL's noise sampling applied to the Cosmos3-Edge video noise produces noise at 0.003 times the typical norm and achieves zero success.

These findings yield two design objectives: a search space whose dimension is independent of $N$, and preservation of the expected noise scale regardless of the actor's output.
\Cref{sec:method} describes how \shortname meets both.

\section{\shortname: Low-Rank Noise Steering}
\label{sec:method}

\begin{figure*}[t]
    \centering
    \includegraphics[width=\textwidth]{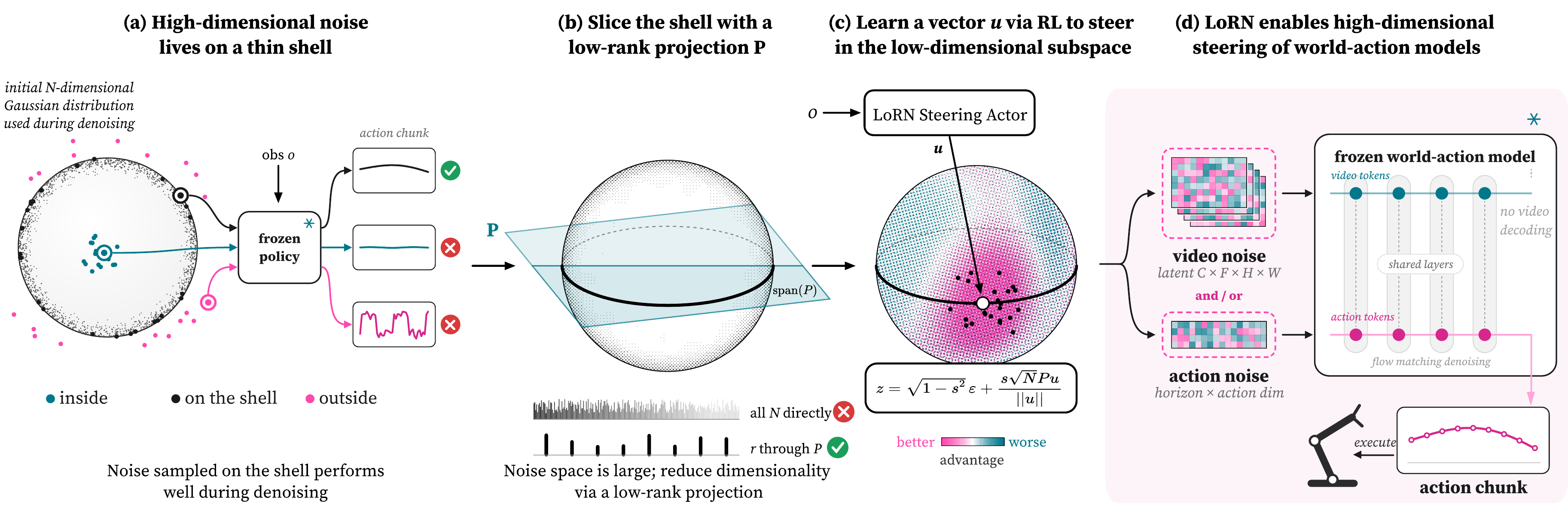}

\caption{\textbf{LoRN steers frozen generative robot policies through a low-dimensional noise subspace.}
\textbf{(a)} High-dimensional Gaussian noise concentrates near a thin shell. We find that moving far from its typical norm can degrade the policy's actions.
\textbf{(b)} A fixed random orthonormal projection $P$ maps $r$ actor coefficients into the full $N$-dimensional noise space, with $r \ll N$.
The plane and its intersection with the sphere illustrate the low-rank subspace.
\textbf{(c)} Given observation $o$, the actor learns coefficients $u$ through reinforcement learning  to produce noise inputs that improve the policy's actions.
The projected output $Pu$ is normalized to $\sqrt{N}$ and mixed with an independent Gaussian sample $\varepsilon$ at strength $s$, using weights that preserve the expected squared noise norm.
The colors illustrate advantage relative to the base policy, implying that our low-rank subspace captures high-advantage regions for policy improvement.
The projection $P$ and pretrained model remain fixed while the steering actor is trained.
\textbf{(d)} This construction enables LoRN to steer video or action noise in a world-action model.
\Cref{alg:lorn} provides more details about LoRN.}    \label{fig:method}
\end{figure*}

The previous section identified two design objectives for steering large policies: a search space whose dimension is independent of the noise dimension $N$, and preservation of the expected squared noise norm regardless of the actor's output, under a fresh Gaussian sample independent of that output.
We construct the steered noise $z$ in four steps that meet these objectives (\cref{sec:construction}), show that the resulting search space contains improving noise (\cref{sec:coverage}), and describe how to choose which noise input of a world-action model to steer (\cref{sec:probe}).
\Cref{alg:lorn} summarizes the procedure.

\subsection{\shortname Construction}
\label{sec:construction}

By the Gaussian annulus theorem (\cref{sec:problem}), samples from a high-dimensional Gaussian, which is used as the source distribution for training flow matching and diffusion policies, lie on a thin shell with a magnitude of $\sqrt{N}$:  $\mathcal{S} = \{z \in \mathbb{R}^N : \|z\| = \sqrt{N}\}$.
We find it useful to view the space of valid initial noise as this sphere. 
A low-rank subspace through the origin intersects the sphere in a lower-dimensional great sphere, analogous to how a plane through the center of a ball cuts its surface in a great circle. 
We depict this method in \cref{fig:method}.
Thus, RL steering with \shortname restricts the coefficients selected by the actor to this low-rank subspace.

Concretely, let $P \in \mathbb{R}^{N \times r}$ be a matrix with orthonormal columns, sampled once from the QR decomposition of a Gaussian matrix and fixed throughout training.
Its columns span an $r$-dimensional subspace, whose intersection with $\mathcal{S}$ is
\begin{equation}
    \mathcal{S}_P = \{\sqrt{N}\, P\hat{u} : \hat{u} \in \mathbb{R}^r,\ \|\hat{u}\| = 1\}.
    \label{eq:slice}
\end{equation}
At each decision, an actor observes $o$ and outputs $r$ coefficients $u$, which select the unit direction $d = Pu/\|u\|$ and hence the point $\sqrt{N} d \in \mathcal{S}_P$.
Each column of $P$ is a full-size noise pattern, so the actor can change every entry of the noise while outputting only $r$ numbers, regardless of $N$.

Rather than using the actor's point directly, we combine it with a Gaussian sample $\varepsilon \sim \mathcal{N}(0, I_N)$:
\begin{equation}
    z = \sqrt{1 - s^2}\, \varepsilon + s \sqrt{N}\, d,
    \label{eq:lorn}
\end{equation}
where $s \in [0, 1]$ is the steering strength.
The Gaussian sample $\varepsilon$ and the actor's unit direction $d$ contribute expected squared norms of $(1-s^2)N$ and $s^2N$, respectively.
Because they are independent and $\varepsilon$ has zero mean, their cross term averages to zero, giving $\mathbb{E}\|z\|^2=N$ for any steering strength $s$.
In high dimensions, the two vectors are also nearly perpendicular and $\|\varepsilon\|$ is close to $\sqrt{N}$, so individual samples of $z$ stay near this norm.
By contrast, using the actor's output directly, as in DSRL, leaves the norm unconstrained, and methods that add a correction to a Gaussian sample push the noise off the sphere.

This Gaussian term also serves as exploration.
It spreads the steered noise around the actor's point, so that each action explores a neighborhood around that point, near the sphere, rather than a single point in $\mathcal{S}_P$.
The strength $s$ controls the width of this neighborhood, much as the entropy temperature controls exploration in SAC: $s = 1$ returns exactly the actor's point, and $s = 0$ returns an unsteered Gaussian sample.
We set $s$ through the length of $u$ with a cap $\alpha$ (\cref{alg:lorn}). \cref{app:strength} discusses this choice.

\begin{algorithm}[!b]
\caption{\shortname}
\label{alg:lorn}
\small
\begin{algorithmic}[1]
\Require Frozen policy $F$, orthonormal $P$, cap $\alpha$, scale $\rho$
\State Initialize SAC actor $\pi_\theta$, critics $Q_\phi$, buffer $\mathcal{D}$
\State For WAMs, select the noise input using \cref{sec:probe}
\For{each decision}
    \State Observe $o$; sample $u \sim \pi_\theta(\cdot \mid o)$
    \State $s \gets \alpha \min\!\left(1, \|u\|/(\rho\sqrt{r})\right)$; $\varepsilon \sim \mathcal{N}(0,I_N)$
    \State $d \gets Pu/\|u\|$ (any unit vector if $u = 0$)
    \State $z \gets \sqrt{1-s^2}\,\varepsilon + s\sqrt{N}\,d$
    \State Execute action chunk $F(o,z)$; observe $R,o'$
    \State Store $(o,u,R,o')$ in $\mathcal{D}$
    \State Update $\theta,\phi$ with SAC using $\mathcal{D}$
\EndFor
\end{algorithmic}
\end{algorithm}

\subsection{Why a Fixed Random Low-Rank Subspace Suffices}
\label{sec:coverage}

We set a projection $P$ at random and keep it fixed for the entirety of RL training.
We investigate whether a randomly chosen low-rank subspace can contain improving noise even when its rank is small relative to the full noise dimension.
We show that, for a fixed fraction of noise inputs that improve on the base policy, the probability that this subspace contains none decreases exponentially with rank, with a bound independent of $N$.
The same projection $P$ is then shared across states and held fixed throughout training.

We analyze the case $s = 1$, where the actor can select any point on $\mathcal{S}_P$.
This idealizes the implemented actor, whose bounded outputs reach $s = 1$ only for directions with $\|u\| \geq \rho\sqrt{r}$ (\cref{app:strength}), and the propositions assume the random projection of \cref{sec:construction} rather than the tiled variants used in some experiments.
We also adopt a spherical-noise model in which unsteered noise is drawn uniformly from $\mathcal{S}$, both for the base policy and for the fallback $u = 0$.
This approximates the Gaussian source by fixing its norm at $\sqrt{N}$, motivated by the concentration of Gaussian norms in high dimensions.
Let $\pi_S$ denote the base policy under this model.

\begin{figure*}[t]
    \centering
    \includegraphics[width=\textwidth]{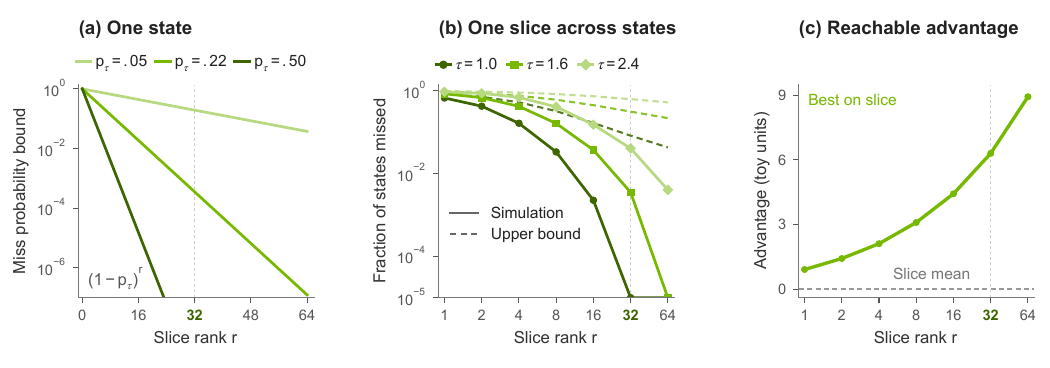}

    \caption{\textbf{A fixed random low-rank subspace misses the good region with probability that decays exponentially in rank.} (a) The bound $(1-p_\tau)^r$ on the probability that a random subspace misses the good region at one state, for three shares $p_\tau$; the dashed line marks rank 32. (b) Simulation in which a single fixed subspace serves 4,000 states, each with its own linear advantage field. Solid: fraction of states whose normalized noise set $\mathcal{S}_P$ contains no noise with advantage at least $\tau$; dashed: the bound $\epsilon_r(\tau)$. (c) The best advantage reachable in $\mathcal{S}_P$ grows with rank, while its average advantage remains zero.}
    \label{fig:coverage}
\end{figure*}

Define
$A(o,z) = Q^{\pi_S}(o,z) - V^{\pi_S}(o)$
as the advantage of using noise $z$ at the current decision and following $\pi_S$ thereafter.
Here, $V^{\pi_S}(o)$ averages $Q^{\pi_S}(o,z)$ over noise sampled uniformly from $\mathcal{S}$.
We use this formulation to define improving regions on the sphere, where for a threshold $\tau > 0$, we can define the improving region $G_\tau(o)$ and its fraction on the sphere $p_\tau(o)$ as
\begin{equation}
\begin{aligned}
    G_\tau(o) &= \{z \in \mathcal{S} : A(o,z) \geq \tau\}, \\
    p_\tau(o) &= \Pr_{z \sim \mathrm{Unif}(\mathcal{S})}
    \big[z \in G_\tau(o)\big].
\end{aligned}
\end{equation}
We say that the subspace \emph{misses} a state $o$ if
$\mathcal{S}_P \cap G_\tau(o) = \varnothing$.

The key observation is that the Gaussian matrix used to construct $P$ has the same column span as $P$.
Scaling each of its columns to norm $\sqrt{N}$ therefore gives $r$ independent uniform samples from $\mathcal{S}$, all lying in $\mathcal{S}_P$.
For the subspace to miss the improving region, all $r$ samples must miss it.
This yields the following guarantees; proofs are in \cref{app:proofs}.

\begin{proposition}[One state]
\label{prop:one}
Fix a state $o$ and sample a uniformly random $r$-dimensional subspace.
Then:
(a) the expected advantage of a uniformly sampled point in $\mathcal{S}_P$ is zero, where the expectation is over both the subspace and the sampled point;
(b) the probability that the subspace misses $G_\tau(o)$ is at most $(1-p_\tau(o))^r$; and
(c) the expected best advantage in $\mathcal{S}_P$ is at least that of the best of $r$ independent uniform samples from $\mathcal{S}$.
\end{proposition}

\begin{proposition}[One fixed subspace across states]
\label{prop:all}
Let $\mu$ be a state distribution independent of $P$, such as the state distribution of the base policy, and define
\[
    \epsilon_r(\tau)
    = \mathbb{E}_{o \sim \mu}
      \big[(1-p_\tau(o))^r\big].
\]
The expected fraction of states under $\mu$ that the subspace misses is at most $\epsilon_r(\tau)$.
Moreover, with probability at least $1-\delta$ over $P$, this fraction is at most $\epsilon_r(\tau)/\delta$, for any $\delta \in (0,1)$.
\end{proposition}

\begin{proposition}[Improvement within the subspace]
\label{prop:improve}
Let $\pi'_P$ select a noise in
$\mathcal{S}_P \cap G_\tau(o)$ whenever this intersection is nonempty, and otherwise use an unsteered sample ($u = 0$).
For every $P$,
\begin{equation}
    J(\pi'_P) - J(\pi_S)
    \geq
    \frac{\tau}{1-\gamma}
    \Big(1-d^{\pi'_P}(M_P)\Big)
    \geq 0,
\end{equation}
where $M_P$ is the set of missed states and
$d^{\pi'_P}$ is the normalized discounted state distribution of $\pi'_P$.
For $\mu = d^{\pi_S}$, the expected one-step advantage, averaged over $P$ and states drawn from $\mu$, is at least
$\tau(1-\epsilon_r(\tau))$.
\end{proposition}

The coverage bound depends on rank and the fraction of improving noise samples, with no explicit dependence on $N$.
For example, if at least 20\% of noise samples at every state have advantage at least $\tau$, a rank-32 subspace misses at most
$0.8^{32} \approx 0.08\%$ of states under the base policy's distribution, in expectation over $P$ (\cref{fig:coverage}).
These guarantees establish the existence of improving choices within a fixed subspace under the spherical-noise model at $s=1$.
How efficiently a trained actor can discover these choices is a separate question, discussed in \cref{app:analysis}.

\subsection{Choosing Which Noise to Steer}
\label{sec:probe}

A world-action model exposes both video and action noise, but these inputs need not have equal influence on the generated actions.
Changing the video noise can also change the action chunk because the model denoises the two inputs jointly.
We therefore choose which input to steer by measuring its effect on the actions before training a steering actor.

For a fixed observation, we first generate a reference action chunk using the model's usual noise inputs.
We then redraw the video noise while holding the action noise fixed, and repeat the experiment by redrawing only the action noise.
For each redraw, we compute the root-mean-square (RMS) difference from the reference action chunk across its time steps and action coordinates.
We average these responses over the probe observations and repeated draws, and select the input with the larger response.
The probe requires only forward evaluations of the frozen model, without rewards or actor training.

\begin{figure}[t]
    \centering
    \includegraphics[width=\linewidth]{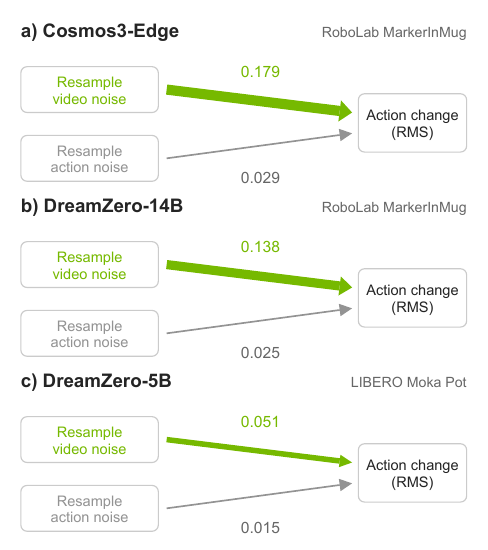}
    \caption{\textbf{Video noise has a larger effect on the generated actions.}
For the same observation, we redraw either video noise (green) or action noise (gray), holding the other input fixed.
Arrow labels show the average root-mean-square (RMS) change from the reference action chunk; arrow widths show relative sensitivity within each model.
Values are in each model's native action units and should be compared within a model.
The larger response to video noise motivates choosing it for steering before actor training.
    }
    \label{fig:probe}
\end{figure}

Surprisingly, resampling video noise produces substantially more variation in the generated actions than resampling action noise: approximately $6.1\times$, $5.5\times$, and $3.3\times$ larger changes on Cosmos3-Edge, DreamZero-14B, and DreamZero-5B, respectively (\cref{fig:probe}).
Although it is associated with visual prediction, video noise may offer a more expressive input for modifying the policy's behavior.
This motivates our choice to steer video noise in the main WAM experiments. 
We evaluate whether this greater influence translates into improved task success in \cref{sec:experiments}.

\section{Experiments}
\label{sec:experiments}

We evaluate \shortname as a way to improve frozen robot policies through noise steering.
We first study video-noise steering across three world-action models, then evaluate action-noise steering on $\pi_0$, $\pi_{0.5}$, and Diffusion Policy.
Controlled experiments examine the roles of the steering rule, projection rank, and initialization, followed by deployment of simulation-trained steering actors on a physical bimanual robot with two base policies, GR00T N1.7 and a fine-tuned $\pi_{0.5}$.
Throughout, the base policy denotes the frozen model without a learned steering actor.

\subsection{Video-Noise Steering Across World-Action Models}
\label{sec:cosmos}
\label{sec:other_wams}

World-action models generate actions jointly with visual predictions, giving the steering actor access to both video and action noise.
The sensitivity probe presented earlier identifies video noise as the more influential input on all three evaluated models.
Here, we test whether learning to steer that input improves task success, and compare it with steering the action noise instead.

We evaluate Cosmos3-Edge and DreamZero-14B~\citep{dreamzero} on four RoboLab tasks~\citep{robolab}: MustardInLeftBin, MarkerInMug, ButterAboveRaisin, and StackYellowOnRed.
We also evaluate DreamZero-5B on the cream-cheese-box and moka-pot tasks from LIBERO~\citep{libero}.
For each model, we compare the base policy, DSRL action-noise steering~\citep{dsrl}, and \shortname applied to either action or video noise.
Comparing the two \shortname variants tests the choice of input under the same steering rule; comparing with DSRL tests against an existing action-noise steering method.
Video steering uses a rank-32 random projection.

Results are means and standard errors over three training seeds.
Cosmos3-Edge uses 240 evaluation episodes per controller; DreamZero-14B uses 120 paired episodes per controller near 200k environment steps; and DreamZero-5B uses its last-three-evaluation summaries.
Base-policy references are the released policies with their own noise samplers, which \shortname recovers at $u=0$ (\cref{app:repro:dreamzero}).
Training budgets, retained checkpoints, and model-specific DSRL implementations are given in \cref{app:repro:cosmos,app:repro:dreamzero}.

\begin{figure*}[t]
  \centering
  \includegraphics[width=\textwidth]{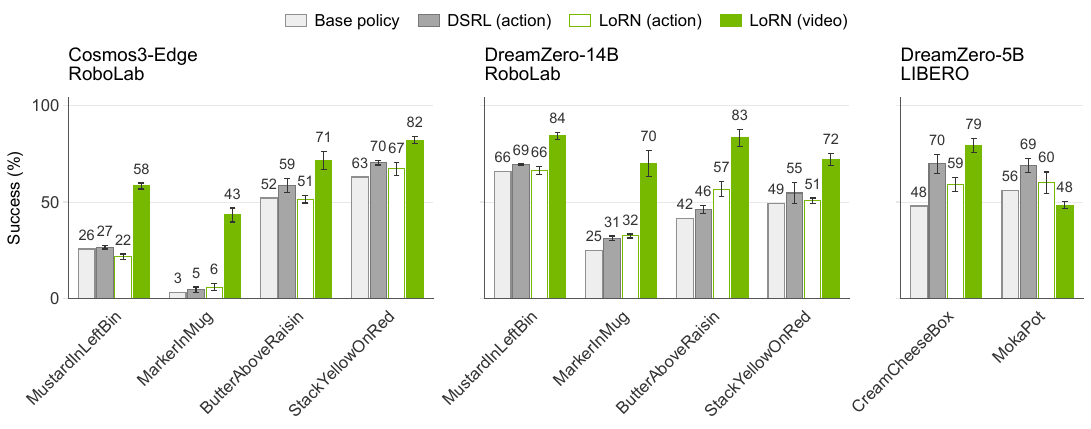}
  \caption{\textbf{Video-noise steering improves success across world-action models.}
Base-policy success, DSRL action-noise steering, and \shortname action- or video-noise steering on RoboLab and LIBERO.
DSRL steers all 256 action-noise entries on Cosmos3-Edge, the stronger of its two forms there (\cref{tab:cosmos}); on DreamZero-14B it tiles a 32-dimensional latent over the chunk, and on DreamZero-5B it replaces the action noise with its latent.
Bars show means; error bars show standard errors over three training seeds.
Base-policy references have no training-seed error bars.
Labels round means to whole percentages.
Video steering achieves the highest mean on nine of ten model--task pairs; DSRL performs best on DreamZero-5B Moka pot.
Evaluation protocols and retained training budgets differ across models and some configurations; see the appendix.}
  \label{fig:wam-success}
\end{figure*}

\Cref{fig:wam-success} shows that video steering achieves the highest mean success on nine of ten model--task pairs, including all four tasks for both Cosmos3-Edge and DreamZero-14B.
Across the ten pairs, average success increases from 43.0\% for the base policy to 50.1\% with DSRL and 69.2\% with \shortname video steering: gains of 7.1 and 26.2 percentage points, respectively, computed from the per-pair means of \cref{tab:cosmos,tab:dz14b,tab:dz5b}.
The DSRL comparator is the stronger of its two Cosmos3-Edge variants; with the tiled variant its average gain is 6.4 points.
These averages weight each model--task pair equally and summarize results under the model-specific evaluation and training protocols.
On Cosmos3-Edge, every video-steering seed exceeds the base policy.

Action-noise steering with \shortname produces smaller gains than video steering, showing that the input choice matters alongside the steering rule.
The exception is DreamZero-5B Moka pot, where DSRL action steering achieves the highest success.
Thus, the sensitivity probe identifies a promising input for learning, but does not guarantee the best learned performance on every task.

\subsection{Action-Noise Steering Across Robot Policies}
\label{sec:action_policies}

We next evaluate action-noise steering beyond world-action models, comparing \shortname with DSRL and LP-DS~\citep{lpds} across $\pi_0$, $\pi_{0.5}$, and Diffusion Policy.
Our evaluation spans twelve tasks across LIBERO, ALOHA, YAM, and Robomimic, with five training seeds per method~\citep{libero,aloha}.
LP-DS uses its released implementation on Robomimic and our adaptations of its steering rule on $\pi_0$ and $\pi_{0.5}$; the Cosmos3-Edge adaptation of \cref{sec:why} is a third, separate construction (\cref{app:repro:action,app:repro:cosmos}).

\begin{figure*}[t]
  \centering
  \includegraphics[width=\textwidth]{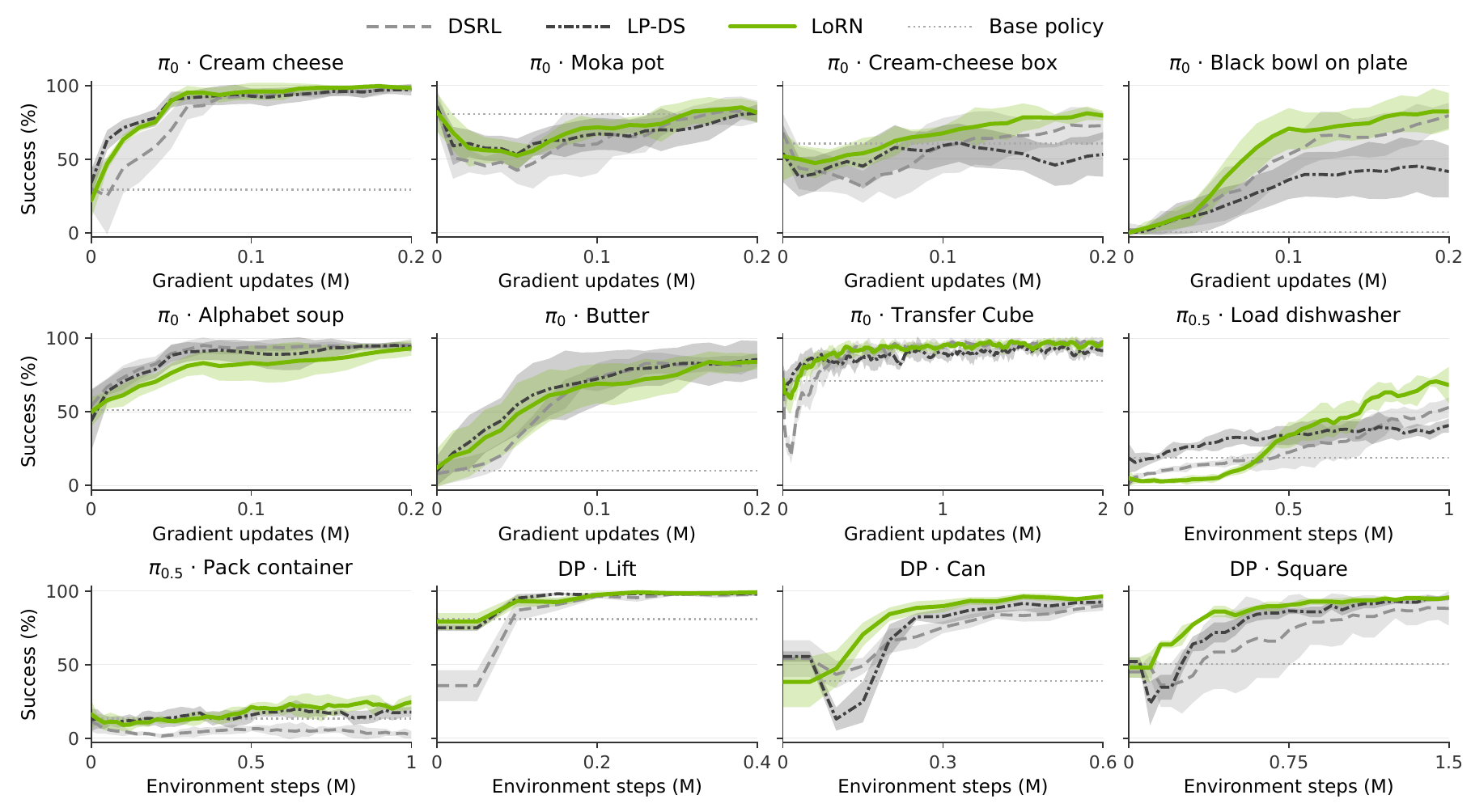}
  \caption{\textbf{Action-noise steering across robot policies.}
Learning curves of \shortname (green), DSRL (gray dashed), and LP-DS (dark gray dash-dotted) on twelve tasks: the mean over five seeds with a band of one standard deviation across seeds.
Dotted horizontal lines mark base-policy success.
The $\pi_0$ tasks use LIBERO and ALOHA, $\pi_{0.5}$ uses YAM, and Diffusion Policy uses Robomimic.
Horizontal axes show gradient updates for $\pi_0$ and environment steps for the other policies, in millions; several panels end before the full training budgets of \cref{tab:repro:runs} to show the phase in which the methods separate, so training budgets should be compared within each task.
The $\pi_0$ and $\pi_{0.5}$ curves are five-evaluation running means of 10- and 16-episode evaluations; the Robomimic curves are raw 200-episode evaluations.
Diffusion Policy uses full-rank \shortname ($r=N=28$).}
  \label{fig:action-policy-curves}
\end{figure*}

As \cref{fig:action-policy-curves} and \cref{tab:action} show, \shortname has the highest early-window mean on nine of the twelve tasks, including ties with LP-DS on cream cheese and Pack container; DSRL leads on alphabet soup, and LP-DS on butter and Load dishwasher.
On $\pi_0$ and Diffusion Policy, its final success is within three points of DSRL on every task.
The strongest final-performance gains appear on $\pi_{0.5}$, where \shortname achieves higher success than both baselines on both tasks.
These results demonstrate that the same steering construction applies across policy families, improving action generation as well as enabling video-noise steering in world-action models.

\subsection{Sim-to-Real Transfer on a Bimanual Robot Station}
\label{sec:real_yam}


\begin{figure*}[t]
  \centering
  \setlength{\tabcolsep}{4pt}
  \begin{tabular}{@{}ccc@{}}
    \includegraphics[width=0.32\textwidth]{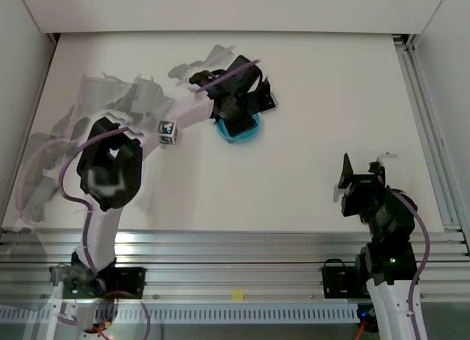} & \includegraphics[width=0.32\textwidth]{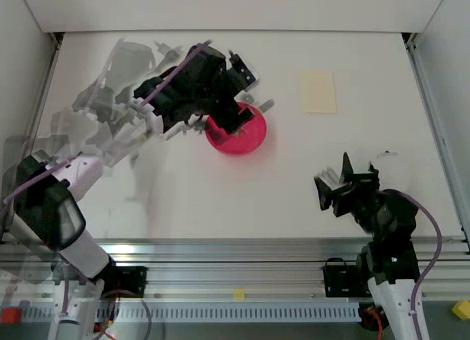} & \includegraphics[width=0.32\textwidth]{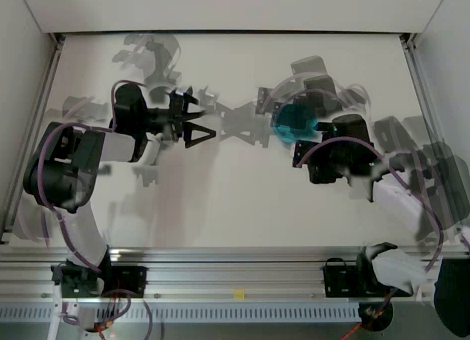}\\
    {\small (a) Cube to bowl} & {\small (b) Marker to cup} & {\small (c) Cube handover}\\
  \end{tabular}
  \caption{\textbf{The three YAM station tasks.} One successful \shortname trial per task from the overhead camera, rendered as in \cref{fig:real_yam}: the initial frame as background, six intermediate arm poses translucent, and the final pose solid. (a) ``Pick up the cube and place it in the bowl.'' (b) ``Pick up the black marker and put it into the red cup.'' (c) ``Pick up the black cube with the left arm, hand it to the right arm, and place it in the blue bowl.'' The steering actors are trained only in an Isaac Lab reconstruction of the station and deployed unchanged.}
  \label{fig:yam_tasks}
\end{figure*}

We test whether steering learned in simulation transfers to a physical robot without additional training on the real robot.
We use a bimanual YAM station on three tasks (\cref{fig:yam_tasks}): cube to bowl and marker to cup with the released GR00T N1.7 base policy, and cube handover, in which the left arm picks up the cube and hands it to the right arm, which places it in a bowl.
For the first two tasks we use the base policy without task-specific fine-tuning; it succeeds on neither task across ten physical trials each.

We reconstruct the YAM setup in Isaac Lab~\cite{mittal2025isaaclab} using a real2sim pipeline with GPT-6 Astra. 
We then train \shortname and DSRL with randomized object poses in this simulation.
Cube positions and orientations vary by $\pm5$\,cm and $\pm45^\circ$, while marker poses vary by $\pm3$\,cm and $\pm29^\circ$.
For each method and task, we select one actor from three simulation-training seeds and deploy it directly on the robot.
We report means and standard errors over three training seeds in simulation and ten trials for real world evaluations.
Actor selection and evaluation details appear in \cref{app:repro:real}.

In simulation, \shortname raises success from 19.2\% to 43.8\% on \texttt{Cube to bowl} and from 3.8\% to 39.6\% on \texttt{Marker to cup}.
DSRL reaches 29.7\% and 28.2\%, respectively (\cref{tab:real_yam}).
On the physical station, \shortname succeeds in 6 of 10 cube trials, compared with 4 of 10 for DSRL.
Both methods succeed in 4 of 10 marker trials.

The released $\pi_{0.5}$ succeeds in none of 120 simulation episodes and none of ten physical trials of the handover task, so we first fine-tune it by supervised fine-tuning (SFT) on 50 scripted simulation demonstrations and then train \shortname on the action noise of the frozen fine-tuned policy (\cref{app:repro:real}).
In simulation, \shortname raises success from 43.6\% to 73.6\%; on the station, the fine-tuned policy places the cube in 3 of 10 trials and \shortname in 8 of 10 (\cref{tab:real_yam}).
Noise steering therefore composes with supervised fine-tuning: the fine-tuned weights stay frozen, and the additional gain comes from the learned noise alone.
These results show that steering learned entirely in simulation can improve real world performance without hardware training.
Notably, the frozen GR00T N1.7 base policy produces successful behavior with noise steering despite failing every unsteered physical trial.
This suggests that the policy already represents behaviors capable of solving these tasks, but its default noise sampling does not reliably elicit them.
\shortname learns to access these behaviors through the noise input, and this control transfers from simulation to the physical robot.
All \shortname physical trials appear in \cref{app:real_trials}.

\begin{table}[!t]
  \centering
  \small
  \caption{\textbf{Simulation-trained steering transfers to the robot station.}
Simulation success (\%): mean $\pm$ s.e. over three training seeds, each evaluated on 240 seeded episodes (120 for the handover task); base policies are evaluated once on the same episodes, and the fine-tuned (SFT) $\pi_{0.5}$ three times, so its spread reflects evaluation rather than training variability.
Robot results are placements in ten trials per policy using one selected actor.
Bold marks the highest value per task; ties are both bold.}
  \label{tab:real_yam}
  \setlength{\tabcolsep}{4pt}
  \renewcommand{\arraystretch}{1.12}
  \begin{tabular}{@{}llcc@{}}
    \toprule
    Task & Policy & Simulation & Robot\\
    \midrule
    Cube to bowl & Base policy & 19.2 & 0/10\\
     & DSRL & 29.7 $\pm$ 10.4 & 4/10\\
     & \shortname & \textbf{43.8 $\pm$ 3.7} & \textbf{6/10}\\
    \midrule
    Marker to cup & Base policy & 3.8 & 0/10\\
     & DSRL & 28.2 $\pm$ 7.6 & \textbf{4/10}\\
     & \shortname & \textbf{39.6 $\pm$ 3.5} & \textbf{4/10}\\
    \midrule
    Cube handover & Base $\pi_{0.5}$ & 0.0 & 0/10\\
     & SFT $\pi_{0.5}$ & 43.6 $\pm$ 2.6 & 3/10\\
     & \shortname on SFT & \textbf{73.6 $\pm$ 3.3} & \textbf{8/10}\\
    \bottomrule
  \end{tabular}
\end{table}

\subsection{Analysis of the Steering Construction}
\label{sec:why}

The previous sections establish that \shortname can improve both video- and action-noise steering in simulation and for the real world.
We now examine the design choices behind these results.
In particular, we investigate how much steering is possible without disrupting the noise norm, what is an ideal low-rank subspace, and how the construction of noise sampling changes the policy before learning.
The first two experiments use Cosmos3-Edge video steering while the latter experiment on noise initialization uses $\pi_0$ action steering.

\paragraph{Steering rule and strength.}
The noise scale probes in \cref{fig:bands} show how changing the noise norm affects a frozen policy.
We next examine how the steering rule affects learning.
On Cosmos3-Edge, we compare four rules for steering video noise on RoboLab's \texttt{MustardInLeftBin} and \texttt{MarkerInMug} tasks.
We use the same rank-32 projection and SAC training procedure, retraining the actor for each rule.

We compare the following methods:
\begin{itemize}
    \item \textbf{\shortname:} projects the actor's output into the full noise space and normalizes it, then mixes it with Gaussian noise while preserving the expected squared noise norm.
    \item \textbf{Additive steering:} adds the mapped actor output to Gaussian noise without adjusting for the resulting increase in noise norm.
    \item \textbf{LP-DS~\citep{lpds}:} adds an observation-dependent offset to Gaussian noise, using an adaptive penalty to constrain the offset's average squared norm.
    \item \textbf{PSS-related control:} applies \shortname's mixture at the fixed strength $s=\sqrt{r/N}\approx0.008$, a dimension-based proxy that gives $r$ of the $N$ noise coordinates to the actor, without the response-based subspace selection or orthogonal-complement reconstruction of PSS~\citep{ni2026pss}.
\end{itemize}
The LP-DS adaptation and the PSS-related control apply the respective steering rules within the same fixed random projection and SAC learner, rather than reproducing the complete methods.
PSS itself selects its subspace by measuring how noise perturbations affect decoded actions, then lets the actor control that subspace while independently sampling Gaussian noise in its orthogonal complement; the control keeps only its dimension-based allocation of actor control, so the comparison isolates the steering rule rather than evaluating the full method.
In contrast, \shortname uses a random subspace and explicitly balances the actor-selected noise with Gaussian noise to preserve the expected squared norm.
We examine the resulting performance differences below; implementation details and training budgets appear in \cref{app:repro:cosmos}.

\begin{figure*}[t]
  \centering
  \includegraphics[width=\textwidth]{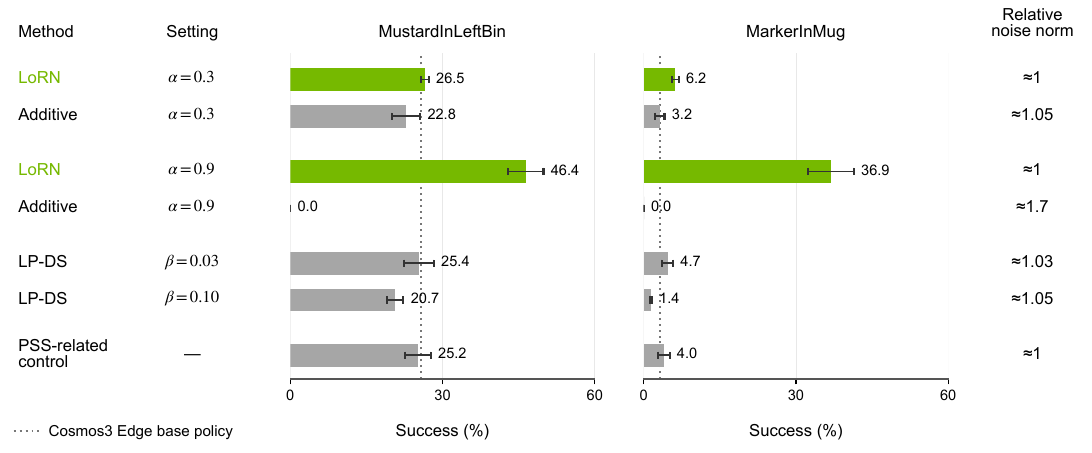}
  \caption{\textbf{\shortname enables stronger steering while maintaining the noise norm.}
Steering-rule comparisons on Cosmos3-Edge using a rank-32 video projection.
Bars show mean success with standard errors over three training seeds; dotted lines mark base-policy success.
Increasing the steering strength $\alpha$ improves \shortname's performance, whereas additive steering increases the noise norm and loses task success.
The LP-DS adaptation~\citep{lpds} and the PSS-related control~\citep{ni2026pss} retain approximately the original noise norm but yield limited gains.
Here, $\alpha$ caps the steering strength while $\beta$ sets the LP-DS correction's target norm limit $\sqrt{\beta N}$.
The relative noise norms are measured in units of $\sqrt{N}$.
Both use our adaptations of the published steering rules within the shared setup, rather than reproductions of the complete methods.}
  \label{fig:steering-rule-controls}
\end{figure*}

\shortname separates the actor's choice of noise from its overall scale (\cref{fig:steering-rule-controls}).
It normalizes the projected actor output and reduces the Gaussian component as steering strength increases, allowing substantial changes to the noise while preserving its expected squared norm.
Additive steering retains the full Gaussian sample, so stronger steering increases the noise norm and leads to failure on both tasks.
Our LP-DS adaptation constrains the size of this additive adjustment, while the PSS-related control assigns only a small fraction of the noise to actor control, leaving the input almost unchanged.
These comparisons suggest that effective steering requires both control over the noise norm and sufficient freedom to modify the noise.

\paragraph{Projection rank.}
\label{sec:projection_rank}
We next examine what rank we should use to steer high-dimensional video noise.
We train Cosmos3-Edge \shortname actors at ranks $4,8,16,32,$ and $256$ on four RoboLab tasks.
Each setting uses three training seeds with independent random projections and is evaluated on 240 episodes.

\begin{figure}[t]
  \centering
  \includegraphics[width=\linewidth]{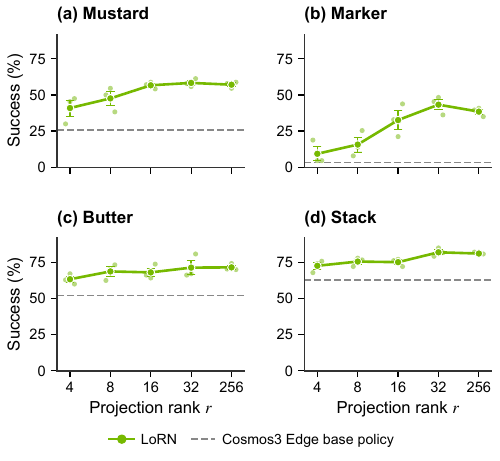}
  \caption{\textbf{A low-rank actor is sufficient for Cosmos3-Edge video steering.}
Success generally improves with projection rank before leveling off; increasing rank from 32 to 256 provides no consistent further gain.
Pale dots show three training seeds, each with an independent random projection and 240 evaluation episodes.
Green lines show means with standard-error bars; dashed lines mark Cosmos3 Edge base-policy success.
Rank settings are evenly spaced, with seed dots offset for visibility.}
  \label{fig:rank_ablation}
\end{figure}

Increasing rank generally improves success before performance levels off (\cref{fig:rank_ablation}).
On Mustard, success rises from $41.0\pm5.5\%$ at rank 4 to $56.7\pm1.3\%$ at rank 16, with little further change at higher ranks.
Marker benefits from additional capacity, improving from 32.6\% at rank 16 to 43.3\% at rank 32.
Across all four tasks, increasing rank eightfold from 32 to 256 provides no consistent further gain.
As such, we use rank 32 as the default for steering Cosmos3-Edge's 506k-dimensional video noise.

Variability across seeds also decreases with rank on certain tasks, suggesting that additional capacity can make steering more reliable.
Because each seed changes both the projection and actor training, this experiment measures their combined effect on learned performance.
This result complements the coverage analysis in \cref{prop:one}, which focuses on the existence of improving noise inputs within the subspace.
\Cref{tab:cosmos_rank} lists these results.

\paragraph{Initialization and noise structure.}
\label{sec:initialization}
Before any learning takes place, initial performance depends on the noise structure that is first supplied.
We examine how the initial noise structure affects starting performance using untrained DSRL and \shortname actors with $\pi_0$ on LIBERO \texttt{CreamCheese} and ALOHA Transfer Cube.
The DSRL implementation repeats a single actor-generated noise vector across the action horizon.
For \shortname, we compare this tiled structure with a random projection that maps the actor's output across the full noise tensor.
We also vary steering strength to examine how retaining more Gaussian noise affects starting performance.

\begin{figure}[t]
  \centering
  \includegraphics[width=0.85\linewidth]{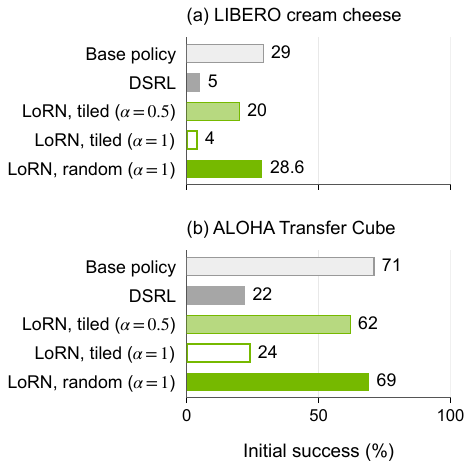}
  \caption{\textbf{Random projections preserve starting performance at full steering strength.}
Success before actor training with frozen $\pi_0$ on two tasks.
Tiled projections repeat noise patterns across the action chunk.
Reducing steering strength from $\alpha=1$ to $0.5$ improves their initial performance, while random-projection \shortname remains near the base policy at $\alpha=1$, without fresh Gaussian noise.
The cream-cheese tiled configuration at $\alpha=0.5$ uses rank 16; all other \shortname configurations use rank 32.
Each bar is the success of the warm-up episodes before the first gradient update, averaged over five training seeds, with latents drawn from a standard normal; their norm saturates the strength mapping of \cref{alg:lorn}, so $s=\alpha$ and the $\alpha=1$ bars use no fresh Gaussian noise. Standard errors are 1.5 to 3.5 points (\cref{app:repro:action}).}
  \label{fig:start}
\end{figure}

DSRL reduces initial success from 29\% to 5\% on cream cheese and from 71\% to 22\% on ALOHA (\cref{fig:start}).
Tiled \shortname at steering strength $\alpha=0.5$ retains more of the base policy's performance, reaching 20\% and 62\%, respectively, but falls to 4\% and 24\% at $\alpha=1$.
This drop implies that tiling is out of distribution for the denoising step of a flow matching policy, but when we set $\alpha=0.5$, we retain performance as we are closer to a Gaussian sample.
\shortname with a random projection instead achieves 28.6\% and 69\% at $\alpha=1$, where the strength saturates at $s=1$, remaining close to the base policy despite using no fresh Gaussian component.
Together, these results suggest that random projections provide a better starting point for steering: they allow the actor to control the entire noise input while retaining the base policy's initial success.

These results highlight initialization as a potential contributor to learning efficiency.
Across our experiments, DSRL exhibits varied starting behavior: some runs begin well below the base policy and recover slowly, while others start near base-policy performance and improve quickly, or begin near maximum success with little room for further improvement.
\shortname's random projection preserves starting performance in these probes while giving the actor a compact space in which to learn.
This combination offers a possible explanation for the rapid improvement observed in several learning curves.
Additional controls on observation dependence, projection structure, rank, and additive steering appear in \cref{app:ablations}.
\Cref{sec:actor_directions} shows how this observation dependence develops during training, from a nearly fixed noise choice to one that varies with the observation (\cref{fig:cloud_training}).

\section{Limitations}
\label{sec:limitations}

Noise steering relies on behaviors that the frozen policy can already produce.
Video noise provides a potentially expressive way to access these behaviors, but greater action variation does not necessarily translate into greater task success.
The sensitivity probe (\cref{fig:probe}) measures how much the actions change, rather than whether those changes help solve the task.
Learning useful control over this variation still requires interaction, and steering cannot directly teach the frozen model a missing skill.

Our analysis shows why a low-rank subspace can contain noise inputs that improve the policy, but does not establish how easily the actor can discover them.
The rank and training required for effective steering therefore remain empirical choices.
Similarly, preserving the native expected energy is a design choice, not a guarantee of the optimal noise scale for every policy: the identity holds for a fresh Gaussian sample independent of the actor's output, which the fixed-draw DreamZero configurations do not satisfy (\cref{app:repro:dreamzero}), and the evaluated Robomimic policies reach their highest success at smaller norms in the amplitude probes of \cref{fig:bands}.

\section{Conclusion}
\label{sec:conclusion}

We introduced \shortname, a method for improving frozen generative robot policies by steering their initial noise through a low-rank subspace.
A small actor controls the noise through a fixed random projection, while normalization and Gaussian mixing allow strong steering at a stable noise scale.
This enables steering of high-dimensional video noise in world-action models, giving the actor access to an input that can substantially influence the generated actions.

Across three world-action models and ten model--task pairs, video steering improves average success by 26.2 percentage points over the base policy, compared with 7.1 points for DSRL action steering.
The same approach supports action-noise steering across several policy families and transfers from simulation to a physical robot without hardware training.
Together, these results establish low-rank noise steering as a practical way to control high-dimensional generative robot policies, including through video noise that provides an expressive input for shaping their actions.

\bibliographystyle{plainnat}
\begingroup
\Urlmuskip=0mu plus 1mu\relax
\bibliography{references}
\endgroup

\onecolumn
\appendix

\section{Theoretical Analysis and Proofs}
\label{app:theory}

This appendix provides the proofs behind our coverage analysis, additional experiments, and implementation details.
The illustrations in \cref{fig:coverage,fig:unrolled} describe the geometry of steering and do not report robot evaluations.

\subsection{Proofs of the Main Propositions}
\label{app:proofs}

The proofs use two properties of the spherical-noise model in \cref{sec:coverage}.
First, $P$ is obtained by QR decomposition of a Gaussian matrix, so its span contains that matrix's normalized columns $g_1,\dots,g_r$.
These are $r$ independent noise vectors distributed uniformly on $\mathcal{S}$.
Second, setting $u=0$ gives $s=0$ and recovers an unsteered sample uniform on $\mathcal{S}$.
Its expected advantage relative to $\pi_S$ is zero by the definition of $V^{\pi_S}$.

\begin{proof}[Proof of \cref{prop:one}]
By rotational invariance, sampling uniformly from $\mathcal{S}_P$ and averaging over the random projection gives a uniform sample on $\mathcal{S}$.
The advantage relative to $\pi_S$ therefore averages to zero, establishing (a).
If $\mathcal{S}_P$ contains no point in $G_\tau(o)$, then all $g_1,\dots,g_r$ must lie outside $G_\tau(o)$.
Their independence gives (b).
Furthermore, $\max_{z\in\mathcal{S}_P}A(o,z)\geq\max_i A(o,g_i)$.
Since the $g_i$ are independent uniform samples from $\mathcal{S}$, this establishes (c).
\end{proof}

\begin{proof}[Proof of \cref{prop:all}]
Because $\mu$ is independent of $P$, we can exchange the expectations over projections and states.
Applying \cref{prop:one}(b) at each state gives the expected fraction of states for which the subspace contains no qualifying noise input.
Markov's inequality gives the corresponding probability bound over projections.
\end{proof}

\begin{proof}[Proof of \cref{prop:improve}]
At a state where $\mathcal{S}_P$ intersects $G_\tau(o)$, the expected advantage of $\pi'_P$ is at least $\tau$.
At a state where it does not, the policy uses an unsteered sample and has expected advantage zero.
Substituting this bound into the performance difference lemma~\citep{kakade2002},
$J(\pi')-J(\pi_S)=\frac{1}{1-\gamma}\mathbb{E}_{o\sim d^{\pi'}}\mathbb{E}_{z\sim\pi'}A(o,z)$,
gives the first statement.
Averaging over $\mu=d^{\pi_S}$ and applying \cref{prop:all} gives the second.
\end{proof}

\subsection{Interpreting the Coverage Bounds}
\label{app:analysis}

\paragraph{Higher rank makes useful noise easier to reach.}
The lower bound on one-step improvement is $\tau(1-\epsilon_r(\tau))$, which can be maximized over $\tau$.
A smaller threshold includes more noise inputs in the qualifying region, while a larger rank reduces the probability of missing that region.
For example, if 20\% of noise inputs on the sphere meet the threshold at every state, rank 32 misses at most approximately 0.8\% of states with probability at least 0.9 over the random projection.
This follows from Markov's inequality and is ten times the expected miss rate in \cref{sec:coverage}.
If 5\% of noise inputs meet a stricter threshold $\tau'$, ranks 32 and 64 give expected one-step improvements of at least approximately $0.81\tau'$ and $0.96\tau'$, respectively.
These examples illustrate the bounds; the corresponding fractions have not been measured on the robot tasks.

\paragraph{Finding useful noise is a separate learning problem.}
The propositions characterize the availability of improving noise inputs within a random subspace.
Learning additionally depends on the critic and on the states visited under the trained actor, whose distribution can depend on $P$.
The state-averaged coverage guarantee applies to a projection-independent state distribution; the projection and rank studies examine learned performance empirically.

\subsection{A Fixed Subspace Across an Episode}
\label{app:episode_geometry}

\begin{figure}[t]
    \centering
    \includegraphics[width=\textwidth]{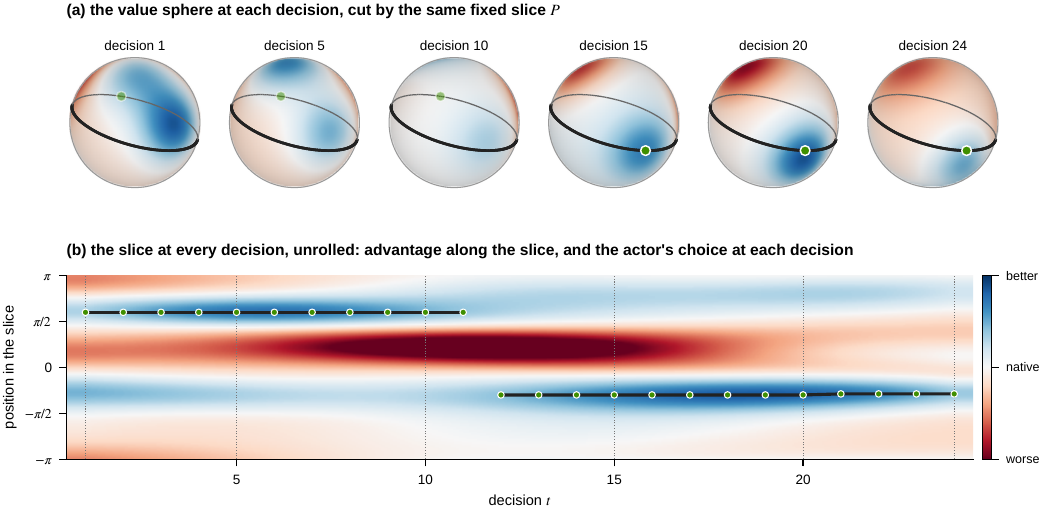}
\caption{\textbf{A fixed low-rank subspace can support different noise choices throughout an episode.}
This schematic illustrates noise steering, rather than measured robot results.
The sphere represents noise inputs at a fixed norm, and its color indicates advantage: expected return relative to the base policy, with blue higher and red lower.
A fixed projection $P$ restricts the actor to the same subspace at every decision; its intersection with the sphere is shown here as a great circle.
(a) Six decisions illustrate how the useful noise inputs change with the observation, while green points mark the actor's choices.
(b) Great circles from 24 decisions are unrolled into columns to show how the actor can track changing high-advantage regions without changing the projection.}
    \label{fig:unrolled}
\end{figure}

The projection remains fixed throughout an episode, so the same coefficient vector $u$ always maps to the same noise input at full steering strength.
Which coefficients are useful can still change with the observation (\cref{fig:unrolled}).
The actor therefore learns in a reparameterized MDP whose actions are coefficients in the fixed subspace.
Changing the projection after training changes the noise associated with those coefficients.

\clearpage

\section{Experimental Setup and Training}
\label{app:reproducibility}

We describe the shared training setup, followed by the environments, policies, and evaluation used for each experiment.
\Cref{app:repro:shared} introduces the shared steering parameterization and reporting conventions.
\Cref{tab:repro:steering,tab:repro:learner,tab:repro:runs} summarize settings across policy families.
The remaining subsections provide the details specific to each experiment.

\subsection{Shared Protocol}
\label{app:repro:shared}\label{app:repro:learner}\label{app:repro:evaluation}\label{app:strength}

\paragraph{\shortname parameterization.}
For actor output $u\in\mathbb{R}^r$ and a fixed projection $P$ with $P^\top P=I_r$, we use
\begin{equation}
\begin{aligned}
 v &= Pu/\|u\|,\qquad
 s=\alpha\min\!\left(1,\frac{\|u\|}{\rho\sqrt r}\right),\\
 z &= \sqrt{1-s^2}\,\varepsilon+s\sqrt N\,v .
\end{aligned}
\label{eq:strength}
\end{equation}
When $u=0$, the implementation returns $\varepsilon$ exactly.
We do not apply a further correction to the norm of each generated sample.
For fresh Gaussian noise $\varepsilon$ independent of $(s,v)$, the construction preserves the conditional expected squared norm, $\mathbb{E}[\|z\|^2\mid s,v]=N$.
This identity does not require every sample to have the same norm.
The DreamZero variants use the fixed-draw convention described in \cref{app:repro:dreamzero}.

The parameter $\alpha$ sets the maximum steering strength, while $\rho$ determines the actor output norm at which this maximum is reached.
Increasing $\rho$ reduces strength for a fixed output $u$.
We distinguish these parameter settings from the strengths measured during training; the empirical sweep appears in \cref{app:strength_results}.

\paragraph{Projection and Gaussian noise.}
We construct $P$ by QR decomposition of a Gaussian matrix, giving fixed orthonormal columns.
For tiled projections, we construct a basis for one row and repeat it over $T$ rows with a factor of $1/\sqrt{T}$ to preserve orthonormality.
The DreamZero-5B video basis repeats across its two generated frames in the same way.
\Cref{tab:repro:steering} summarizes the noise dimension, projection rank and structure, and steering strength for each policy family.
The Gaussian component is resampled at each decision except in the two DreamZero variants.

\begin{table}[h]
\centering\small
\setlength{\tabcolsep}{4pt}
\renewcommand{\arraystretch}{1.12}
\caption{\textbf{A small actor controls high-dimensional noise across policy families.}
Settings for \shortname, which maps $r$ actor outputs through a fixed projection into an $N$-dimensional noise input while keeping the base-policy weights frozen.
$\alpha$ sets the maximum steering strength $s$: $s=0$ uses the original noise and $s=1$ uses only the normalized projected actor output.
Random bases have orthonormal columns; tiled bases repeat across action-chunk steps; frame-constant bases repeat across generated video frames.
Output-dependent strengths use $s=\alpha\min(1,\|u\|/(\rho\sqrt{r}))$ with $\rho=0.3$ and actor output $u$.
The two YAM station actors instead use fixed $s=1$.
Diffusion Policy uses $r=N$, corresponding to a random rotation of the full noise space.
Projection seeds are separate from the learner seeds.}
\label{tab:repro:steering}
\begin{tabular}{@{}lrrlll@{}}
\toprule
Policy / noise input & $N$ & $r$ & Basis & $\alpha$ & Projection seed(s)\\
\midrule
Cosmos3-Edge video & 506{,}880 & 32 & Random & 1 & 1234--1236, one per learner seed\\
DreamZero-14B video & 112{,}640 & 32 & Random & 1 & 1234; action 4321\\
DreamZero-5B video & 25{,}600 & 32 & Frame-constant & 1 & 1234; action 4321\\
$\pi_0$ LIBERO / ALOHA & 1{,}600 & 16 & Tiled & 0.5 & 4321, all learner seeds\\
$\pi_{0.5}$ & 960 & 32 & Random & 1 & 4321, all learner seeds\\
Diffusion Policy & 28 & 28 & Random & 0.5 & 4321, all learner seeds\\
GR00T N1.7 action (YAM station) & 1{,}280 & 32 & Random & 1 (fixed $s$) & 4321, all learner seeds\\
Fine-tuned $\pi_{0.5}$ action (YAM station) & 420 & 32 & Random & 1 (fixed $s$) & 4321, all learner seeds\\
\bottomrule
\end{tabular}
\end{table}

\paragraph{Learner.}
Steering actors use SAC~\citep{sac}; \cref{tab:repro:learner} summarizes the implementations, architectures, and hyperparameters.
The model-specific subsections describe the observations supplied to each actor and critic.
Actor and critic inputs are shared, without privileged state information, except on the handover task, where the critics additionally receive the simulator's object poses during training; the deployed actor never does (\cref{app:repro:real}).
The frozen policy receives the language instruction, but the steering actor does not receive it as a separate input.
Each replay transition corresponds to one executed action chunk.
Rewards and bootstrapping conventions follow the task setups described below.

\begin{table}[tb]
\centering\small
\setlength{\tabcolsep}{3pt}
\renewcommand{\arraystretch}{1.12}
\caption{\textbf{Soft Actor-Critic settings for training noise-steering actors.}
The generative robot policy remains frozen while the actor and critics learn from environment interaction.
Each steering decision produces one action chunk and one replay transition; replay capacities count these transitions.
Robomimic denotes the Diffusion Policy experiments, and GR00T N1.7 denotes the YAM station experiments.
Hidden layers are listed as number of layers $\times$ width.
Learning rates are ordered actor / critic / entropy temperature; critic aggregation is ordered temporal-difference target / actor objective.
$d_u$ is the actor output dimension, and the temperature weights the entropy term in SAC.
Slash-separated task settings mean LIBERO / ALOHA for $\pi_0$ and Cosmos3-Edge / DreamZero-14B in their shared column.
All optimizers use Adam with target-network update rate 0.005; ``Auto'' denotes learned temperature and dashes denote inapplicable settings.
The YAM station's DSRL baseline uses a transformer actor rather than the listed \shortname MLP.}
\label{tab:repro:learner}
\begin{tabular}{@{}>{\raggedright\arraybackslash}p{.16\textwidth}*{6}{>{\raggedright\arraybackslash}p{\dimexpr.14\textwidth-2\tabcolsep\relax}}@{}}
\toprule
Setting & $\pi_0$ & Robomimic & Cosmos / DreamZero-14B & DreamZero-5B & $\pi_{0.5}$ & GR00T N1.7 (YAM station)\\
\midrule
Implementation & jaxrl2 SAC & SB3 DSRL-NA & PyTorch SAC & PyTorch SAC & PyTorch SAC & PyTorch SAC\\
Hidden layers & $3\times128$ & $3\times2048$ & $3\times512$ & $3\times128$ & $2\times256$ & $2\times256$\\
Critics; aggregation & 10; mean / mean & 2; min / min & 10; mean / mean & 10; mean / mean & 4; mean of 2 / min & 4; mean of 2 / min\\
Learning rates & $10^{-4}$ / $3\!\times\!10^{-4}$ / $3\!\times\!10^{-4}$ & $3\!\times\!10^{-4}$ for all & $10^{-4}$ / $3\!\times\!10^{-4}$ / $3\!\times\!10^{-4}$ & $10^{-4}$ / $3\!\times\!10^{-4}$ / $3\!\times\!10^{-4}$ & $5\!\times\!10^{-5}$ / $10^{-4}$ / $10^{-4}$ & $5\!\times\!10^{-5}$ / $10^{-4}$ / --\\
Batch size & 256 & 256 & 768 & 512 & 64 & 64\\
Replay capacity & 25k / 150k & $10^7$ & 1{,}900 (recent decisions only) & 48{,}000 & 200{,}000 & 200{,}000\\
Discount per decision & $0.999^{20}$ / $0.999^{50}$ & 0.99; Square 0.999 & 0.99 & $0.999^{16}$ & $0.995^{30}$ & $0.9975^{40}$\\
Entropy target & $-d_u/2$ / 0 & 0 & 0 & $-16$ & 0 & --\\
Temperature & Auto; init. 1 & Auto & Softplus; init. 0.008, floor 0.005 & Softplus; init. 1 & Log; automatic & Fixed 0.05\\
Actor output bound & $\pm1$ / $\pm2$ & $\pm1.5$ & $\pm1$ & $\pm1$ & $\pm1$ & $\pm1$\\
Warmup & 500 / 1{,}000 decisions & 1{,}501 / 2{,}001 vector steps$^a$ & 48k env. steps & 19{,}200 env. steps & 450 env. steps & 450 env. steps\\
Updates per collection unit & 20 / decision & 20 / vector step; Lift 30 & 2 / decision & 2 / decision & 2 / decision & 2 / decision\\
Parallel train envs. & 1 & 4 & 12 / 8 & 1 & 8 & 8\\
\bottomrule
\end{tabular}
\par\vspace{3pt}\begin{minipage}{\textwidth}\footnotesize
Slash-separated settings refer to LIBERO / ALOHA or Cosmos3-Edge / DreamZero-14B, as appropriate.
$^a$Robomimic warmup counts vector steps, each collecting from four environments, and uses frozen-policy data.
The YAM station uses a fixed temperature and a transformer actor for DSRL's action-noise outputs.

\end{minipage}
\end{table}

\paragraph{Frozen policies and success.}
All base-policy weights remain frozen, and methods within a comparison use the same checkpoint.
Evaluation success follows the simulator's task predicate at the end of an episode.
An evaluation timeout counts as failure.

\paragraph{Reported checkpoints and estimates.}
Training progress is recorded as gradient updates for $\pi_0$ and environment steps for the other policy families.
We compare learning curves within each task.
\Cref{tab:repro:runs} summarizes the training budgets and the evaluations used for reported success.
Results use the specified checkpoints or trailing windows; the physical deployment separately selects one actor per method and task from the simulation-trained seeds.
DSRL and \shortname use sampled actor outputs during evaluation, except on DreamZero-5B, which uses deterministic outputs.
LP-DS uses the actor mode for $\pi_0$ and the Polyak-averaged actor in the released Robomimic implementation.

\begin{table}[tb]
\centering\small
\setlength{\tabcolsep}{4pt}
\renewcommand{\arraystretch}{1.12}
\caption{\textbf{Training budgets and the evaluations used to summarize success.}
Each row identifies a policy family, its nominal training budget, the number of independent training seeds, and how success is calculated for each seed.
Reported means and standard errors are computed across these seed-level estimates.
Environment steps count executed low-level actions; $\pi_0$ progress instead counts gradient updates, and Diffusion Policy uses wall-clock budgets.
A trailing-window estimate averages distinct checkpoint evaluations, rather than evaluating one final controller on all pooled episodes.
Paired evaluations share initial conditions across methods.
Some Cosmos3-Edge runs use earlier checkpoints near 0.9--1.0M steps, and two DreamZero-5B video runs use checkpoints near 0.46--0.48M steps.
The YAM station rows describe simulation evaluation; physical deployment separately evaluates one simulation-selected actor per method and task on ten trials.}
\label{tab:repro:runs}
\begin{tabularx}{\textwidth}{@{}l>{\raggedright\arraybackslash}Xc>{\raggedright\arraybackslash}X@{}}
\toprule
Policy & Training budget & Seeds & Reported evaluation per seed\\
\midrule
Cosmos3-Edge & 1.5M env. steps; Marker 1.0M & 3 & 240 episodes at the specified checkpoint\\
DreamZero-14B & 200k env. steps & 3 & 120 paired episodes near the end of training\\
DreamZero-5B & 614{,}400 env. steps & 3 & Mean of the last three 25-episode evaluations\\
$\pi_0$ & 500k gradient updates (LIBERO); 3M (ALOHA) & 5 & Mean of the last ten smoothed evaluations (\cref{app:repro:action})\\
$\pi_{0.5}$ & 1M env. steps & 5 & Mean of the last ten evaluations\\
Diffusion Policy & 12 h (Lift/Can); 30 h (Square) & 5 & Mean of the last ten 200-episode evaluations\\
GR00T N1.7 (YAM station) & 1M env. steps & 3 & 240 simulation episodes at the final checkpoint\\
Fine-tuned $\pi_{0.5}$ (YAM station) & 3M env. steps & 3 & 120 simulation episodes at the final checkpoint\\
\bottomrule
\end{tabularx}
\end{table}

\paragraph{Tasks, episodes, and rewards.}
\Cref{tab:repro:tasks} lists, for every policy family, the task identifiers, episode limits, reset randomization, reward, and in-training evaluation cadence.
Projection seeds are in \cref{tab:repro:steering}; learner seeds are numbered from 0, with the seed count $K$ of \cref{tab:repro:runs}.

\begin{table}[tb]
\centering\small
\setlength{\tabcolsep}{3.5pt}
\renewcommand{\arraystretch}{1.12}
\caption{\textbf{Task configuration.} Episode limits count environment steps. In-training evaluation gives the number of episodes and the cadence at which they run, in gradient updates for $\pi_0$ and environment steps otherwise. Rewards are per decision unless stated.}
\label{tab:repro:tasks}
\begin{tabular}{@{}>{\raggedright\arraybackslash}p{.13\textwidth}>{\raggedright\arraybackslash}p{.27\textwidth}>{\raggedright\arraybackslash}p{.08\textwidth}>{\raggedright\arraybackslash}p{.17\textwidth}>{\raggedright\arraybackslash}p{.13\textwidth}>{\raggedright\arraybackslash}p{.12\textwidth}@{}}
\toprule
Policy & Tasks & Episode limit & Reset randomization & Reward & In-training evaluation\\
\midrule
Cosmos3-Edge & RoboLab MustardInLeftBin, MarkerInMug, ButterAboveRaisin, StackYellowOnRed & 448, 600, 576, 896 & Object positions uniform within $\pm5$\,cm, 0.25\,m separation floor & $+1$ on the success transition, 0 otherwise; time limits bootstrap & 24 episodes / 15k env.\ steps\\
DreamZero-14B & The same four RoboLab tasks & Same & Same & Same & 16--20 episodes / 20--40k env.\ steps\\
DreamZero-5B & LIBERO-90 cream-cheese box (\texttt{LIVING\_ROOM\_SCENE1}, index 47) and moka pot (\texttt{KITCHEN\_SCENE3}, index 19) & 600 & LIBERO initial states & Same & 25 episodes / 38{,}400 env.\ steps\\
$\pi_0$ & LIBERO-90 cream cheese (57), moka pot (19), cream-cheese box (47), black bowl (9), alphabet soup (55), butter (56); ALOHA Transfer Cube & 400 & LIBERO: the benchmark's 50 initial states; ALOHA: default environment resets & $-1$ per decision, 0 on success; only success disables bootstrapping & 10 episodes / 10k gradient updates\\
$\pi_{0.5}$ & YAM Load dishwasher; Pack container & 300; 450 & Dishwasher: recorded scene; Pack: bottles $\pm6$--8\,cm, box $\pm2$\,cm & Terminal: 0 on success, $-1$ on failure or timeout & 16 episodes / 18k env.\ steps\\
Diffusion Policy & Robomimic Lift, Can, Square & 300, 300, 400 & Standard resets & Sparse reward minus one, summed over the chunk; time limits terminal & 200 episodes / ${\approx}$48k env.\ steps (64k on Square)\\
GR00T N1.7 (YAM station) & Cube to bowl; Marker to cup & 900 & Cube $\pm5$\,cm and $\pm45^\circ$; marker $\pm3$\,cm and $\pm29^\circ$ & \S\ref{app:repro:real} & 32 episodes / 50k env.\ steps\\
Fine-tuned $\pi_{0.5}$ (YAM station) & Cube handover & 1{,}350 & Cube $x\in[0.28,0.44]$, $y\in[-0.21,-0.05]$\,m, yaw $\pm45^\circ$; bowl $x\in[0.36,0.50]$, $y\in[-0.56,-0.40]$\,m & Staged: 0.02 hold, 0.05 lift, 0.15 handover, 0.30 place; $+1$ on success & 120 episodes at the final checkpoint\\
\bottomrule
\end{tabular}
\end{table}

\paragraph{Success estimates and uncertainty.}
Success rate is the fraction of evaluation episodes that succeed.
We first calculate an estimate for each training seed, then report the mean and standard error across seeds, with standard error equal to the sample standard deviation divided by $\sqrt{K}$.
For a single checkpoint evaluated on $M$ episodes, the binomial standard error $\sqrt{\hat{p}(1-\hat{p})/M}$ instead measures episode-sampling uncertainty.
These two sources of uncertainty should not be interpreted interchangeably.
Table values use the aggregation specified in \cref{tab:repro:runs}; for $\pi_0$, whose evaluations contain ten episodes, the last-ten mean is taken on the five-evaluation running mean defined in \cref{app:repro:action}.

\paragraph{Compute.}
Experiments use NVIDIA L40 and L40S GPUs.
Runs on $\pi_0$, Robomimic, DreamZero-5B, $\pi_{0.5}$, and the YAM station tasks use one GPU each.
Cosmos3-Edge and DreamZero-14B use two GPUs, with the model and simulator on separate devices.
Run times are reported below for each policy family.

\subsection{Amplitude and Sensitivity Probes}
\label{app:repro:probes}

\paragraph{Amplitude probes.}
We multiply the frozen policy's Gaussian noise by an amplitude $c$ and compare success across amplitudes on matched initial states (\cref{fig:bands}).
The grid contains 17 amplitudes from 0.02 to 2.3, except for DreamZero-14B, which uses 14 amplitudes up to 2.0.
Cosmos3-Edge is evaluated on Mustard using 100 trajectories per amplitude: 50 initial contexts with two samples each.
DreamZero-14B is evaluated on Marker using 100 episodes per amplitude, with the same seeded initial conditions across amplitudes.
The $\pi_0$ probe uses Cream cheese with 100 paired episodes per amplitude, and Diffusion Policy uses Can with 200.
We define the empirical tolerated range in \cref{fig:bands} as the tested amplitudes whose success is no more than one standard error below the fresh-noise reference.
This range summarizes the measured grid; it is not a confidence interval for an underlying threshold.
Squared-norm ratios, where reported, are the squares of the corresponding amplitude ratios.

\paragraph{Noise-input sensitivity probe.}
The sensitivity probe follows \cref{sec:probe} and reports the root-mean-square change in decoded actions in each model's own action units (\cref{fig:probe}).
For Cosmos3-Edge and DreamZero-14B, we use 12 Marker states from two base-policy episodes and four fresh noise draws per state.
DreamZero-5B is probed on LIBERO Moka pot.
For DreamZero-14B, the ratios of action changes from resampling video versus action noise are 7.5, 4.3, 5.5, and 3.0 on Mustard, Butter, Marker, and Stack, respectively.
The probe requires no reward and selects video noise for all three evaluated models.

\subsection{Cosmos3-Edge Training and Evaluation}
\label{app:repro:cosmos}

\paragraph{Environment and policy.}
We use the \texttt{Cosmos3-\allowbreak Edge-\allowbreak Policy-\allowbreak DROID} checkpoint on \texttt{MustardInLeftBin}, \texttt{MarkerInMug}, \texttt{ButterAboveRaisin}, and \texttt{StackYellowOnRed} in RoboLab.
The model generates eight video frames per decision, and we execute 32 actions.
Object poses are randomized within each task, and methods are evaluated on the same initial placements.

\paragraph{Steering and training.}
Video \shortname uses a rank-32 random projection of the 506{,}880 noise entries at $\alpha=1$ and $\rho=0.3$, giving a measured strength of approximately 0.98.
Action \shortname uses a rank-32 projection of the 256 action-noise entries.
The DSRL variants either repeat eight actor outputs across the action horizon or control all 256 entries directly.
The actor and critics receive a trainable encoding of the previous decision's conditioning-frame latent and eight joint/gripper coordinates.
The reward is $+1$ on success and zero otherwise; time-limit transitions retain bootstrapping.

\paragraph{Evaluation.}
We evaluate three training seeds per configuration on 240 paired episodes.
Per-seed success rates are rounded to one decimal before the mean and standard error across seeds are computed, which can move a reported mean by up to 0.05 points relative to the pooled counts.
The nominal budgets appear in \cref{tab:repro:runs}.
Some Stack runs use available checkpoints near 1.0M steps, and one Butter video-steering run uses its 0.9M-step checkpoint; the corresponding curves had remained flat for at least 400k steps.
These are the checkpoints used for the reported comparison.

\paragraph{Steering-rule controls.}
The controls in \cref{tab:cosmos}b use the same rank-32 video projection and learner, with a separately trained actor for each steering rule.
Replacement steering directly maps actor outputs to noise as $z=cPu$.
We compare unscaled replacement, fixed rescaling by $\sqrt{N/r}$, and a scale calibrated at initialization to match the Gaussian expected squared norm.
The calibrated scale remains fixed during training.
These are projected-replacement controls rather than reproductions of DSRL's complete method.
The additive control adds the normalized projected actor output to the full Gaussian sample,
\begin{equation}
    z_{\mathrm{add}} = \varepsilon + \frac{s}{\sqrt{1-s^2}}\sqrt{N}\,\frac{Pu}{\|u\|},
    \label{eq:additive_control}
\end{equation}
which, for the same $u$, $\varepsilon$, and $s<1$, equals \shortname's noise divided by $\sqrt{1-s^2}$.
It therefore preserves the resulting vector's orientation but raises its expected squared norm to $N/(1-s^2)$ under independent fresh Gaussian noise.
The same maximum steering strength does not imply the same measured strength after training.
At $\alpha=0.9$, Mustard runs average $s=0.88$ with \shortname and $s=0.82$ with additive steering.
Our LP-DS adaptation adds a projected actor output to Gaussian noise and uses a dual variable to constrain its mean squared norm,
\[
 z=\varepsilon+\sqrt{N/r}\,Pu,
 \qquad \operatorname{mean}(u^2)\le b,
\]
where $b\in\{0.03,0.10\}$ is the target mean squared coefficient magnitude.
In table labels we denote this setting by $\beta=b$.
The corresponding correction has target root-mean-square norm $\sqrt{bN}$.
The constraint is optimized using projected dual ascent.
The PSS-related control instead uses \shortname's mixture at fixed strength $s=\sqrt{r/N}=0.0080$.
It matches a dimension-based allocation of actor control but does not implement PSS's response-based projection selection or its Gaussian orthogonal-complement reconstruction.
These comparisons therefore evaluate adapted mechanisms under a shared projection, rather than the complete LP-DS or PSS methods.
The lower-strength and norm-constrained controls use available checkpoints after their learning curves level off, between 0.86M and 1.0M steps; the $\alpha=0.9$ comparison uses the full 1.5M-step budget.

\paragraph{Rank, projection, and steering strength.}
The rank sweep compares $r\in\{4,8,16,32,256\}$ using three training seeds with independent random projections (\cref{tab:cosmos_rank}).
Stack evaluations use checkpoints near 1.0M steps, consistent with the main comparison.
The projection study uses nine actors across five projections, with three independently trained actors on each of two projections and one on each of the remaining three.
Projection-swap tests change only $P$, keeping the actor and evaluation initial states fixed.
The strength study varies $\rho\in\{0.3,0.6,1\}$ under a shared projection and training seed, together with an $\alpha=0.5$ comparison (\cref{app:strength_results}).

\paragraph{Compute.}
Each run uses two GPUs and takes 36--64 h.

\subsection{DreamZero Training and Evaluation}
\label{app:repro:dreamzero}

Both DreamZero variants reuse the sampler's original Gaussian draw as $\varepsilon$ in \cref{eq:strength}.
Setting $u=0$ therefore recovers the released sampler exactly.
Because this draw is fixed, the fresh-Gaussian expected squared-norm identity in \cref{app:repro:shared} does not apply.
The DreamZero base-policy references are therefore the released policies with their fixed draws, evaluated on the same paired initial states as the trained actors, as for every other model.

\subsubsection{DreamZero-14B on RoboLab}

\paragraph{Environment and policy.}
We use the \texttt{DreamZero-\allowbreak DROID} checkpoint on the same four RoboLab tasks as Cosmos3-Edge.
The model predicts 24-step action chunks using its default denoising schedule, and we execute each chunk in full.
Video and action \shortname use rank-32 random projections at $\alpha=1$.
DSRL repeats its 32-dimensional actor output across the action horizon.
The observation encoders, reward, and learner follow the Cosmos3-Edge setup.

\paragraph{Training and evaluation.}
We train three seeds per configuration for 200k environment steps and evaluate the resulting checkpoints on 120 paired episodes per seed.
Initial states are shared across methods, including the base-policy reference.
Each run uses two GPUs and takes approximately 20 h.

\subsubsection{DreamZero-5B on LIBERO}

\paragraph{Environment and policy.}
We use the RLinf 5B LIBERO SFT checkpoint (\texttt{RLinf-\allowbreak DreamZero-\allowbreak WAN2.2-\allowbreak 5B-\allowbreak LIBERO-\allowbreak SFT-\allowbreak Step26000}) on Cream-cheese box and Moka pot from LIBERO-90.
The model executes 16-action chunks using eight DiT evaluations per decision.
Video \shortname uses a rank-32 basis repeated across the generated frames at $\alpha=1$.
Action \shortname uses a tiled rank-16 basis at $\alpha=0.5$.
The actor and critics receive image features from trainable convolutional encoders together with joint/gripper coordinates.
Rewards and bootstrapping follow the Cosmos3-Edge setup.

\paragraph{Training and evaluation.}
We train three seeds per configuration with a nominal budget of approximately 0.61M environment steps.
Results average each seed's last three 25-episode evaluations, using deterministic actor outputs.
Two Moka-pot video-steering runs ended earlier, at approximately 0.46M and 0.48M steps, and use their last available evaluations.
The base-policy reference uses the released fixed noise draw on 25 episodes.
Each run uses one GPU and takes 20--27 h.

\subsection{Action-Policy Training and Evaluation}
\label{app:repro:action}

\paragraph{Early and final success.}
The appendix's early-window mean is each seed's average success over the window $[0,B]$.
For $\pi_0$, whose evaluations contain only 10 episodes, we first smooth the curve with a trailing five-evaluation running mean, in which each evaluation is averaged with the four preceding ones, or with all preceding ones during the first four evaluations; the smoothed curve is resampled on a grid of 10k gradient updates by holding the latest evaluation, integrated by the trapezoid rule over the grid from the first evaluation, at step 0, to $B$, and divided by $B$.
For Robomimic, the same trapezoid is taken over the raw 200-episode evaluations on a 50k-step grid.
For $\pi_{0.5}$, whose 16-episode evaluations are evenly spaced about 18k environment steps apart, the early-window mean is the plain average of the raw evaluations up to $B$.
The endpoint $B$ is 200k gradient updates for LIBERO, 1M gradient updates for ALOHA, 500k environment steps for Lift and both $\pi_{0.5}$ tasks, and 1M environment steps for Can and Square.
Seeds that do not reach $B$ have no early-window estimate.
Final success is the mean of each seed's last ten evaluations (\cref{tab:repro:runs}); for $\pi_0$ these are the last ten values of the same five-evaluation running mean, so each seed's estimate is a weighted mean of its last 14 raw evaluations.
We distinguish the released LP-DS implementation used on Robomimic from our adaptations to $\pi_0$ and $\pi_{0.5}$.
The $\pi_0$ and $\pi_{0.5}$ learning curves of \cref{fig:action-policy-curves} are drawn with the five-evaluation running mean; the Robomimic curves are raw.

\paragraph{LP-DS variants.}
We use the released LP-DS implementation on Robomimic and evaluate several adaptations for $\pi_0$ and $\pi_{0.5}$ (\cref{tab:lpds_variants}).
The variant comparison covers the original five $\pi_0$ tasks; alphabet soup and butter were added later and run with the reported variant only.
We report the best-performing tested variant for each policy family.
On Robomimic, we use the released per-task targets.
On $\pi_0$, the strongest variant places the release's residual steering rule within the DSRL training setup: a 32-dimensional offset repeated across the action chunk, a per-coordinate mean-square target of 0.35, an output bound of 5, and DSRL's latent critic.
This variant reproduces the cream-cheese performance reported by LP-DS.
A target of 0.10 and the two Algorithm 1 variants perform less well on most tested tasks.
On $\pi_{0.5}$, the release-rule adaptation fails with an output bound of 1 and remains below the base policy with a bound of 5.
Our Algorithm 1 adaptation with $\delta=0.35$ on the repeated row improves both tasks and is the reported variant.
Using the release's per-coordinate target in that adaptation leads to failure on most seeds.
An initial adaptation with a stochastic offset was stopped early because LP-DS specifies a deterministic offset.

\begin{table}[tb]
\centering\small
\setlength{\tabcolsep}{4pt}
\renewcommand{\arraystretch}{1.12}
\caption{\textbf{LP-DS variants used to compare action-noise steering across policy families.}
LP-DS adds an observation-dependent offset to Gaussian noise and constrains its magnitude.
We use its released implementation on Diffusion Policy and compare adaptations of the released steering rule and the paper's Algorithm 1 on $\pi_0$ and $\pi_{0.5}$.
A ``broadcast row'' is a 32-dimensional offset repeated across action-chunk steps; ``box'' denotes the bound on each offset coordinate.
The target $\delta$ constrains the offset magnitude, with row-wise and per-coordinate definitions distinguished in the variant descriptions.
Slash-separated values give final success (\%), calculated as each seed's mean over its last ten evaluations (smoothed for $\pi_0$ as in \cref{app:repro:action}) and then averaged over five seeds unless a row specifies one seed.
Task order is $\pi_0$: LIBERO Cream cheese / Moka pot / Cream-cheese box / Black bowl / ALOHA Transfer Cube; $\pi_{0.5}$: YAM Load dishwasher / Pack container; Diffusion Policy: Robomimic Lift / Can / Square.
``Reported'' marks the variant selected for the main comparisons; rows without numerical success describe stopped runs.}
\label{tab:lpds_variants}
\begin{tabular}{@{}l>{\raggedright\arraybackslash}p{.40\textwidth}>{\raggedright\arraybackslash}p{.33\textwidth}c@{}}
\toprule
Policy & Variant & Final success (\%) & Reported\\
\midrule
\multirow{4}{*}{$\pi_0$} & Release rule in the DSRL recipe: per-coordinate target 0.35, box 5 & 97.0 / 78.0 / 64.7 / 64.5 / 94.1 & yes\\
 & Same, target 0.10 & 92.8 (cream cheese only) & \\
 & Algorithm 1, $\delta=0.35$ on the broadcast row (one seed) & 62 / 92 / 58 / 0 / 85 & \\
 & Algorithm 1 with the release's per-coordinate target, 0.35 per coordinate or 11.2 on the 32-d row (one seed) & 100 / 88 / 54 / 1 / 81 & \\
\midrule
\multirow{4}{*}{$\pi_{0.5}$} & Algorithm 1, $\delta=0.35$ on the broadcast row & 39.8 / 18.9 & yes\\
 & Algorithm 1 with the release's per-coordinate target, 11.2 on the 32-d row & 13.4 / 3.6 & \\
 & Release rule, deterministic shift, box 5 & 5.5 / 5.4 & \\
 & Release rule, deterministic shift, box 1 & at zero from 75k steps; stopped at 0.6M & \\
\midrule
Diffusion Policy & Released implementation: $\delta=0.35$ on Can and Square, 0.05 on Lift & 99.2 / 98.0 / 98.3 & yes\\
\bottomrule
\end{tabular}
\end{table}

\subsubsection{\texorpdfstring{$\pi_0$}{pi0} on LIBERO and ALOHA}

\paragraph{Environment and policy.}
We use openpi's \texttt{pi0\_libero} checkpoint on six LIBERO tasks and \texttt{pi0\_aloha\_sim} on ALOHA Transfer Cube.
Both use their default flow sampler and predict 50 actions per decision; we execute 20 on LIBERO and 50 on ALOHA.
\shortname controls the 1{,}600 action-noise entries through a tiled rank-16 projection at $\alpha=0.5$.
Actor and critic encoders process a camera image together with proprioception.
The reward is $-1$ per decision and zero on success; only success disables bootstrapping.
Training uses jaxrl2 SAC with the budgets and five-seed evaluation protocol in \cref{tab:repro:runs}.
A run takes 5--6 h on LIBERO or 28 h on ALOHA using one GPU.

\paragraph{Baselines and ablations.}
The LP-DS adaptation repeats a 32-dimensional offset across the action horizon, using the release's per-coordinate mean-square target of 0.35 and output bound of 5.
Its critic is trained directly on $(o,\Delta)$ without distillation, and collection and evaluation use the actor mode.
The Cream-cheese target ablation uses 0.10 instead of 0.35.
For additive steering, we compare maximum strengths $\alpha$ of 0.5, 0.75, and 0.9; the actor's output norm stays above $\rho\sqrt{r}$ throughout these runs, so the delivered strength equals $\alpha$.
The observation-free control zeros the actor inputs while keeping the critic observations unchanged.
Tiled and random projections are compared at matched rank and strength, and the rank study uses tiled projections of rank 4, 8, 16, and 32 at strength 0.5.
Initialization probes are the warm-up round before any learning: latents are drawn from a standard normal, whose norm exceeds $\rho\sqrt{r}$, so the strength mapping saturates at $s=\alpha$ under the corresponding projections.

\subsubsection{\texorpdfstring{$\pi_{0.5}$}{pi0.5} on the YAM Tasks}

\paragraph{Environment and policy.}
We use the LeRobot \texttt{pi05-MolmoAct2-YAM} checkpoint on Load dishwasher and Pack container.
The policy uses ten flow steps and executes 30-action chunks in full.
Load dishwasher uses a recorded scene, while Pack container randomizes object placements.
\shortname controls the 960 action-noise entries through a rank-32 random projection at $\alpha=1$.
Actor and critics receive frozen DINOv2-small features from three cameras and 14 joint coordinates.
The terminal reward is zero on success and $-1$ on failure or timeout.
Training uses PyTorch SAC with five seeds and a 1M-environment-step budget.
Each run uses one GPU and takes 20--24 h.

\paragraph{LP-DS adaptation.}
Our adaptation follows Algorithm 1 of LP-DS with a deterministic 32-dimensional offset repeated across the action horizon.
It trains an action critic through temporal-difference learning and distills a noise critic from the stored decoded actions.
A dual variable targets a squared offset-row norm of $\delta=0.35$.
The implementation conditions the offset on the observation, bootstraps from the stored successor chunk, and has no target actor.

\subsubsection{Diffusion Policy on Robomimic}

\paragraph{Environment and policy.}
We use the DPPO Robomimic MLP checkpoints on Lift, Can, and Square, with eight DDIM steps and four-action chunks.
\shortname controls all 28 action-noise entries through a rank-28 random rotation at $\alpha=0.5$.
Actor and critics receive the standard low-dimensional robot and object observations.
Rewards sum the sparse reward minus one over the action chunk, and time limits are terminal.
We train five seeds per method using the wall-clock budgets in \cref{tab:repro:runs}, reaching approximately 1.2M steps on Lift, 1.6M on Can, and 3.6--4.0M on Square.
Each run uses one GPU.

\paragraph{Baselines and ablations.}
DSRL uses the SB3 DSRL-NA implementation, which distills a noise critic from an action critic and initializes replay with frozen-policy data.
LP-DS uses the released implementation with a deterministic offset, a zero-initialized output layer, and a Polyak-averaged actor.
Its output bound is 1.5, with a mean-square offset target of 0.35 on Can and Square and 0.05 on Lift.
The Lift target follows the released configuration rather than the 0.10 listed in the paper.
The additive and observation-free controls follow the same definitions as for $\pi_0$.

\subsection{Sim-to-Real Transfer on the YAM Station}
\label{app:repro:real}

\paragraph{Environment and tasks.}
The YAM station has two arms, an overhead camera, and a wrist camera on each arm.
Its Isaac Lab reconstruction matches the scene assets, arm kinematics, and calibrated camera views.
Cube to bowl requires placing a cube in a bowl, Marker to cup requires placing a whiteboard marker in the target container, and Cube handover requires the left arm to pick up the cube and hand it to the right arm, which places it in a bowl.
We randomize object poses during simulation training.
Training uses a red cube and red cup; the physical YAM station uses a black cube and a pink bowl as the marker target, with task instructions unchanged.
\Cref{fig:yam_tasks} shows the three tasks on the physical station.

\paragraph{Policy and steering.}
For Cube to bowl and Marker to cup we use the released GR00T N1.7 policy without task-specific fine-tuning.
It receives the three camera views, a task instruction, and a 32-dimensional state describing both arms.
The policy predicts 40-step chunks, with the joint and gripper commands executed in full.
Both methods steer the 1{,}280 action-noise entries.
\shortname uses a rank-32 random projection at fixed strength $s=1$.
DSRL generates the full action-noise tensor using a transformer actor across the chunk steps.

\paragraph{Observations and rewards.}
The steering actor and critics receive frozen DINOv2-base features from the overhead camera and the robot state.
Wrist images are supplied only to the frozen policy.
The \shortname actor and critics use two-layer MLPs; shared SAC settings appear in \cref{tab:repro:learner}.
Cube to bowl uses a terminal reward of 0 on success and $-1$ on failure or timeout, while Marker to cup pays staged rewards of 0.1 for contact, 0.2 for grasping, 0.2 for lifting, 0.1 for holding the marker over the cup, and 0.3 for placing it inside, each once per episode, plus 1 on success.
A trial succeeds when the released object remains inside the target container through the final second.
Timeouts count as failure.

\paragraph{Training and evaluation.}
We train three simulation seeds per method and task for 1M environment steps.
Each final actor is evaluated on 240 episodes with initial poses shared across methods.
The base policy is evaluated on the same episodes using fresh Gaussian noise.
For physical deployment, we select one actor per method and task using simulation performance and run ten trials on the YAM station without further training.
The base policy also runs ten trials per task.
Objects are placed by hand near their nominal initial poses, and placement success is judged from the camera recordings.
All \shortname physical trials appear in \cref{fig:real_yam}.
Each simulation-training run uses one GPU.

\paragraph{Cube handover with a fine-tuned $\pi_{0.5}$.}
The released $\pi_{0.5}$ completed none of our handover evaluations, so we first fine-tune its action expert, with the vision-language backbone frozen, on 50 scripted simulation demonstrations for three epochs (4{,}884 updates, learning rate $2.5\times10^{-5}$, effective batch 32).
\shortname then steers the action noise of the frozen fine-tuned policy through a rank-32 random projection over the 30-step, 14-channel action block at fixed strength $s=1$; the padded channels keep their Gaussian noise.
The actor and critics receive frozen DINOv2 features of the three cameras and the robot state, and the critics additionally receive the simulator's object poses.
The reward adds staged bonuses for holding, lifting, handing over, and placing the cube, and $+1$ on success.
We train three seeds for 3M environment steps on eight parallel scenes with the learner of \cref{tab:repro:learner} and a discount of $0.9975^{30}$ per decision, and evaluate each final actor on 120 episodes with randomized cube and bowl positions (\cref{tab:repro:tasks}); the fine-tuned policy is evaluated three times on the same episodes.
On the station, the fine-tuned policy and one simulation-selected \shortname actor each run ten trials; a trial succeeds when the cube rests in the bowl at the end.

\clearpage
\section{Additional Results and Ablations}
\label{app:main}\label{app:noise_steering_tables}

The following results supplement the main experiments.
Evaluation protocols and implementation details appear in \cref{app:reproducibility}.

\begin{figure*}[t]
    \centering
    \includegraphics[width=\textwidth]{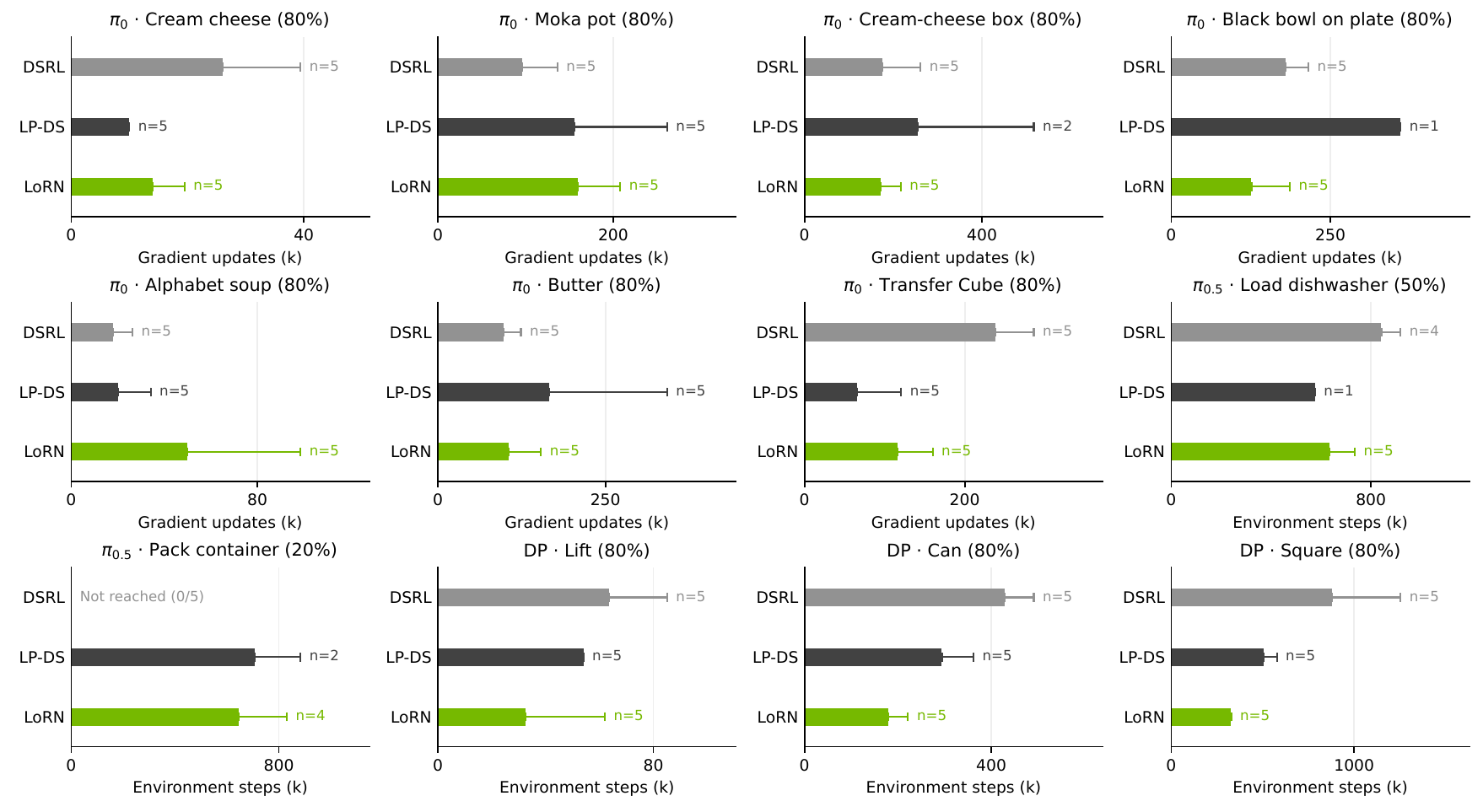}
\caption{\textbf{Sample efficiency of action-noise steering across twelve robot tasks.}
Horizontal bars show the mean training steps to sustained target success for DSRL (gray), LP-DS (dark gray), and \shortname (green); lower is better.
Whiskers show one sample standard deviation across the seeds contributing to each bar.
The success target is 80\% on ten tasks, 50\% on Load dishwasher, and 20\% on Pack container, as indicated in the panel labels.
Means and standard deviations are computed over attaining seeds only, out of five training seeds per method; the label on each bar gives the number of attaining seeds $n$, and bars with different $n$ are not directly comparable.
Horizontal axes use thousands of gradient updates for $\pi_0$ and thousands of environment steps for the other policies, with a separate scale for each task; attainment times use discrete evaluation checkpoints without interpolation or smoothing.
\shortname reaches the target first on six of the twelve tasks, including all three Robomimic tasks; DSRL is first on Moka pot, Alphabet soup, and Butter, and LP-DS on Cream cheese, Transfer Cube, and Load dishwasher.}
\label{fig:action-sample-efficiency}
\end{figure*}

\subsection{Cosmos3-Edge Video Steering}
\label{app:cosmos}

\paragraph{Video steering improves all four Cosmos3-Edge tasks.}
\Cref{tab:cosmos} gives the numbers behind \cref{fig:wam-success,fig:steering-rule-controls}.
Video steering improves on the base policy on every task and training seed, by approximately 33 percentage points on Mustard, 40 on Marker, and 19 each on Butter and Stack; the stronger DSRL action-steering variant gains at most 7.5 points.
With initial placements paired across methods, individual video-steering seeds improve by 30--35 points on Mustard and 33--45 on Marker, with episode-level standard errors of approximately four points.
Action and video \shortname share the construction and the actor output dimension, so their gap isolates the choice of input.

\paragraph{Keeping the norm stable makes stronger steering possible.}
Additive steering at $\alpha=0.9$ and calibrated replacement move outside the tolerated range of \cref{fig:bands} and reach zero success.
The LP-DS adaptation and the PSS-related control keep the norm near its native value but leave the actor little control, and gain little; \shortname keeps the norm and gives the actor control, and gains most.
Settings and implementation differences are in \cref{app:repro:cosmos}.

\paragraph{An eightfold increase beyond rank 32 adds little.}
\Cref{tab:cosmos_rank} lists the rank sweep of \cref{fig:rank_ablation}.
Mustard rises from 41.0\% at rank 4 to 56.7\% at rank 16, Marker from 32.6\% at rank 16 to 43.3\% at rank 32, and rank 256 adds no consistent gain on any task at the evaluated budgets.
The protocol, including the 1.0M-step Marker budget, is in \cref{app:repro:cosmos}.

\paragraph{Different projections work, but the actor learns to use its own.}
Across nine actors trained on five random projections, success ranges from 54\% to 63\% on Mustard and from 36\% to 50\% on Marker, so every tested projection supports improvement.
Replacing the projection of a trained actor without retraining drops Mustard success from 57.9\% to 29.2\% and 35.8\% for two replacement projections, and Marker success from 45.4\% to 3.8\% and 10.0\%.
Useful noise is thus available in many subspaces, but the learned coefficients are specific to the training projection, which must stay fixed at deployment.

\begin{table}[tb]
  \centering
  \small
  \setlength{\tabcolsep}{4pt}
  \renewcommand{\arraystretch}{1.12}
\caption{\textbf{Video steering improves Cosmos3-Edge, and its noise construction supports stronger actor control.}
Success (\%) on four RoboLab tasks: MustardInLeftBin (Mustard), MarkerInMug (Marker), ButterAboveRaisin (Butter), and StackYellowOnRed (Stack).
Values are mean $\pm$ standard error over three training seeds, each evaluated on 240 episodes; the base policy is the frozen model without steering.
(a) DSRL replaces action noise with either eight outputs repeated across the horizon or 256 outputs controlling every entry.
\shortname maps 32 actor outputs into either action or video noise; video steering achieves the highest mean on all four tasks.
(b) Actors are retrained under different steering rules using the same rank-32 video projection.
$\alpha$ limits steering strength, and relative norm is measured against the typical Gaussian norm $\sqrt{N}$ for noise dimension $N$.
Additive steering retains the full Gaussian sample; calibrated replacement uses a fixed scale chosen at initialization.
The LP-DS adaptation constrains an additive offset, with $\beta$ setting its target mean squared coefficient magnitude.
The PSS-related control uses \shortname's mixture at fixed strength $\sqrt{r/N}$, rather than the complete PSS method.
The main video configuration delivers strength $s\approx0.98$ and is distinct from the lower-strength controls.}
  \label{tab:cosmos}

  \begin{tabularx}{\textwidth}{@{}>{\raggedright\arraybackslash}p{0.30\textwidth}>{\centering\arraybackslash}p{0.085\textwidth}*{4}{>{\centering\arraybackslash}X}@{}}
    \toprule
    \multicolumn{6}{@{}l}{\textbf{(a) Which noise to steer}}\\[3pt]
    Method & Input & Mustard & Marker & Butter & Stack\\
    \midrule
    Base policy & -- & 25.8 & 3.3 & 52.1 & 62.9\\
    DSRL, tiled 8-d latent & Action & 26.5 $\pm$ 2.2 & 3.6 $\pm$ 0.6 & 55.0 $\pm$ 0.8 & 68.6 $\pm$ 1.0\\
    DSRL, all 256 entries & Action & 26.7 $\pm$ 1.0 & 4.6 $\pm$ 1.5 & 58.6 $\pm$ 3.5 & 70.4 $\pm$ 1.2\\
    \shortname, rank 32 & Action & 21.7 $\pm$ 1.5 & 6.1 $\pm$ 1.7 & 51.4 $\pm$ 1.9 & 67.2 $\pm$ 3.4\\
    \shortname, rank 32 & Video & \textbf{58.3 $\pm$ 1.6} & \textbf{43.3 $\pm$ 3.6} & \textbf{71.4 $\pm$ 4.7} & \textbf{82.0 $\pm$ 1.7}\\
    \bottomrule
  \end{tabularx}

  \vspace{7pt}
  \begin{tabularx}{\textwidth}{@{}>{\raggedright\arraybackslash}p{0.36\textwidth}*{4}{>{\centering\arraybackslash}X}@{}}
    \toprule
    \multicolumn{5}{@{}l}{\textbf{(b) Controls with the rank-32 video projection}}\\[3pt]
    Steering rule & Strength $\alpha$ & Relative norm & Mustard & Marker\\
    \midrule
    \shortname & 0.3 & $\approx1$ & 26.5 $\pm$ 0.8 & 6.2 $\pm$ 0.7\\
    Additive & 0.3 & $\approx1.05$ & 22.8 $\pm$ 2.8 & 3.2 $\pm$ 0.9\\[2pt]
    \shortname & 0.9 & $\approx1$ & 46.4 $\pm$ 3.5 & 36.9 $\pm$ 4.5\\
    Additive & 0.9 & $\approx1.7$ & 0.0 $\pm$ 0.0 & 0.0 $\pm$ 0.0\\
    \midrule
    Calibrated replacement & -- & Drifts & 0.0 $\pm$ 0.0 & 0.0 $\pm$ 0.0\\
    LP-DS, $\beta=0.03$ & -- & $\approx1.03$ & 25.4 $\pm$ 2.9 & 4.7 $\pm$ 1.1\\
    LP-DS, $\beta=0.10$ & -- & $\approx1.05$ & 20.7 $\pm$ 1.6 & 1.4 $\pm$ 0.2\\
    PSS-related control & $\sqrt{r/N}$ & $\approx1$ & 25.2 $\pm$ 2.5 & 4.0 $\pm$ 1.1\\
    \bottomrule
  \end{tabularx}
  \vspace{3pt}
\parbox{\textwidth}{\footnotesize Panel (b) uses our LP-DS adaptation and a PSS-related control within the shared random projection, rather than reproductions of the complete methods.
For LP-DS, $\beta$ sets the target mean squared coefficient magnitude; the corresponding correction has target root-mean-square norm $\sqrt{\beta N}$.
The PSS-related control uses \shortname's mixture at fixed $s=\sqrt{r/N}$, without PSS's response-based subspace selection or orthogonal-complement reconstruction.
Calibrated replacement uses a constant scale while the actor's output norm changes during training.
The rank sweep appears in \cref{tab:cosmos_rank}.}
\end{table}

\begin{table}[tb]
    \centering
    \small
    \setlength{\tabcolsep}{4pt}
    \renewcommand{\arraystretch}{1.12}
\caption{\textbf{Rank 32 is sufficient for strong Cosmos3-Edge video steering across four tasks.}
Success (\%) after training \shortname with different numbers $r$ of actor outputs, mapped through fixed random projections into the model's 506{,}880-dimensional video noise.
Task names abbreviate RoboLab MustardInLeftBin (Mustard), MarkerInMug (Marker), ButterAboveRaisin (Butter), and StackYellowOnRed (Stack).
The numbered columns are projection ranks; ``Base'' is the frozen policy using its original Gaussian noise without steering.
Values are mean $\pm$ standard error across three independently trained actors with independent projections, each evaluated on 240 episodes.
Success generally increases at lower ranks, while increasing rank from 32 to 256 provides no consistent further gain.}
    \label{tab:cosmos_rank}
    \begin{tabular}{lcccccc}
        \toprule
        Task & Base & 4 & 8 & 16 & 32 & 256 \\
        \midrule
        Mustard & 25.8 & 41.0 $\pm$ 5.5 & 47.6 $\pm$ 4.9 & 56.7 $\pm$ 1.3 & 58.3 $\pm$ 1.6 & 57.1 $\pm$ 1.3 \\
        Marker  & 3.3  & 9.3 $\pm$ 4.7  & 15.7 $\pm$ 5.1 & 32.6 $\pm$ 6.5 & 43.3 $\pm$ 3.6 & 38.5 $\pm$ 1.8 \\
        Butter  & 52.1 & 63.3 $\pm$ 2.1 & 68.7 $\pm$ 3.2 & 68.1 $\pm$ 2.9 & 71.4 $\pm$ 4.7 & 71.5 $\pm$ 1.3 \\
        Stack   & 62.9 & 72.6 $\pm$ 2.4 & 75.6 $\pm$ 1.8 & 75.1 $\pm$ 1.5 & 82.0 $\pm$ 1.7 & 81.1 $\pm$ 0.5 \\
        \bottomrule
    \end{tabular}
\end{table}

\subsection{DreamZero Video Steering}
\label{app:dz14b}

\paragraph{Video steering also improves all four tasks on DreamZero-14B.}
\Cref{tab:dz14b} provides the numerical results for DreamZero-14B in \cref{fig:wam-success}.
Each training seed is evaluated at a checkpoint near 200k environment steps and compared with the released frozen policy on the same paired initial states (\cref{app:repro:dreamzero}).
Video steering exceeds this baseline for every seed on all four tasks.
With video noise redrawn at every decision instead of the released fixed draw, the frozen policy scores 77.5, 41.7, 51.7, and 30.0 on Mustard, Marker, Butter, and Stack, and every video-steering seed also exceeds these values.
The advantage of video steering is particularly clear on Mustard and Marker, where it gains 18 and 45 points over the base policy, while action steering gains at most 8 points and does not exceed the fresh-draw reference.
Together with Cosmos3-Edge, this extends the result across two world-action models on the same tasks.
Both models expose action noise to steering, yet the larger gains come from controlling their video input.
This supports the view that video noise is an actionable part of a world-action model's control interface, even when the task objective concerns only the executed actions.
The sensitivity probe motivates this choice before training; the success evaluations establish that the actor can use that influence to improve behavior.

\begin{table}[tb]
    \centering
    \small
    \setlength{\tabcolsep}{4pt}
    \renewcommand{\arraystretch}{1.12}
\caption{\textbf{Video steering improves DreamZero-14B on all four RoboLab tasks.}
Success (\%) on MustardInLeftBin (Mustard), MarkerInMug (Marker), ButterAboveRaisin (Butter), and StackYellowOnRed (Stack).
Each trained configuration uses three seeds with checkpoints near 200k environment steps; values are mean $\pm$ standard error across seeds evaluated on 120 episodes with matched initial conditions across methods.
The base policy is the released sampler with its fixed draw, which the trained actors recover at $u=0$.
DSRL replaces action noise with a 32-dimensional output repeated across the 24-step action horizon.
\shortname uses 32 actor outputs projected into either action or video noise at maximum strength $\alpha=1$.
Video steering achieves the highest mean success on every task; bold entries mark the highest mean per task.}
    \label{tab:dz14b}
    \begin{tabular}{llcccc}
        \toprule
        Method & Input & Mustard & Marker & Butter & Stack \\
        \midrule
        Base policy & -- & 65.8 & 25.0 & 41.7 & 49.2 \\
        DSRL, tiled 32-d latent & Action & 69.4 $\pm$ 0.3 & 31.4 $\pm$ 1.1 & 46.1 $\pm$ 2.0 & 54.7 $\pm$ 5.4 \\
        \shortname, rank 32 & Action & 66.4 $\pm$ 2.0 & 32.5 $\pm$ 1.0 & 56.7 $\pm$ 3.8 & 50.6 $\pm$ 1.5 \\
        \shortname, rank 32 & Video & \textbf{84.2 $\pm$ 1.9} & \textbf{70.0 $\pm$ 6.6} & \textbf{83.1 $\pm$ 4.3} & \textbf{71.9 $\pm$ 3.1} \\
        \bottomrule
    \end{tabular}
\end{table}

\paragraph{The most useful noise input can depend on the task.}
\Cref{tab:dz5b} reports DreamZero-5B results on two LIBERO-90 tasks.
On Cream-cheese box, video steering improves success from a base-policy reference of 48\% to approximately 79\%, compared with 70\% for DSRL.
On Moka pot, DSRL action steering achieves approximately 69\%, while video steering reaches 48\%, compared with 56\% for the base policy.
The task-dependent results distinguish sensitivity from useful control: the probe identifies video noise as a promising input (\cref{fig:probe}), while task success also depends on what the actor learns.
Greater action variation gives the actor more ways to modify behavior, but those changes need not align equally well with every task's reward.
These results motivate using the probe as a starting point for input selection, followed by task-based evaluation of the learned controller.
They also suggest that the value of an input should be assessed for the task and training setting, rather than treated as a fixed ranking of video and action noise.
Results use the evaluation summaries in \cref{app:repro:dreamzero}, including the available checkpoints for two Moka-pot runs that ended before the nominal budget.

\begin{table}[tb]
    \centering
    \small
    \setlength{\tabcolsep}{4pt}
    \renewcommand{\arraystretch}{1.12}
\caption{\textbf{Video and action steering provide different benefits on DreamZero-5B.}
Success (\%) on the LIBERO-90 Cream-cheese-box and Moka-pot tasks, with the generative policy frozen throughout training.
DSRL replaces action noise; action \shortname uses a tiled rank-16 projection at steering strength limit $\alpha=0.5$; video \shortname uses a rank-32 basis repeated across generated frames at $\alpha=1$.
Each trained result first averages the last three 25-episode evaluations of one seed, then reports mean $\pm$ standard error across three training seeds.
The nominal budget is approximately 0.61M environment steps, with two Moka-pot video runs ending near 0.46M and 0.48M steps.
The base policy uses the released fixed noise draw and is evaluated on 25 episodes.
Video steering has the highest mean on Cream-cheese box, while DSRL action steering has the highest mean on Moka pot; bold marks the highest mean per task.}
    \label{tab:dz5b}
    \begin{tabular}{llcc}
        \toprule
        Method & Input & Cream-cheese box & Moka pot \\
        \midrule
        Base policy & -- & 48.0 & 56.0 \\
        DSRL & Action & 69.8 $\pm$ 4.9 & \textbf{68.9 $\pm$ 3.6} \\
        \shortname, tiled rank 16 & Action & 59.1 $\pm$ 3.6 & 60.0 $\pm$ 5.6 \\
        \shortname, rank 32 & Video & \textbf{79.1 $\pm$ 3.6} & 48.4 $\pm$ 1.9 \\
        \bottomrule
    \end{tabular}
\end{table}

\subsection{Steering Strength and Noise Structure}
\label{app:strength_results}

\begin{figure}[tb]
    \centering
    \includegraphics{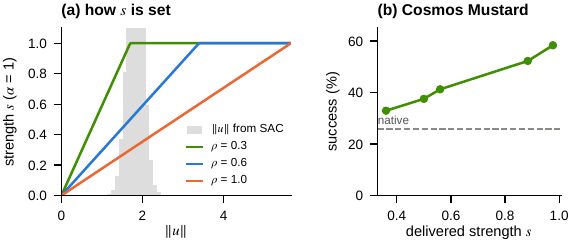}
\caption{\textbf{Steering strength controls how much of the noise is selected by the actor.}
Cosmos3-Edge video steering maps actor output $u\in\mathbb{R}^{r}$ through a fixed rank-$r$ projection and mixes the normalized result with Gaussian noise.
Strength $s=0$ uses only Gaussian noise, while $s=1$ uses only the actor-selected noise.
(a) The mapping $s=\alpha\min(1,\|u\|/(\rho\sqrt{r}))$ is shown at $\alpha=1$ for three values of $\rho$; larger $\rho$ gives weaker steering for the same output norm.
The gray histogram is a tanh-Gaussian fit to recorded actor output norms.
Under this fit, mean strengths are 0.98, 0.54, and 0.32 for $\rho=0.3$, 0.6, and 1, respectively.
(b) Success on RoboLab MustardInLeftBin increases over the tested range of measured steering strengths, while the construction maintains the expected squared noise norm.}
    \label{fig:strength}
\end{figure}

\paragraph{Steering strength stays stable as the actor learns.}
The parameter $\rho$ controls how the actor's output norm determines steering strength (\cref{app:strength}).
In the measured runs, the resulting strength changes little during training: approximately 0.35 at $\rho=1$ and 0.56 at $\rho=0.6$ from initialization to the final checkpoint (\cref{fig:strength}a); the tanh-Gaussian fit in the caption of \cref{fig:strength} gives 0.32 and 0.54.
Mustard success increases with measured steering strength over the tested range, up to approximately $s=0.98$ (\cref{fig:strength}b).
The stable strengths help interpret what changes during learning.
At these settings, the actor improves its noise choices without a large shift in how much Gaussian noise is retained.
Learning therefore need not operate by gradually increasing steering strength; it can improve how the available control is used.
The success sweep complements this observation by showing that giving the actor more control is beneficial on Mustard when the noise scale is maintained.

\paragraph{Retaining Gaussian noise can preserve a useful starting point.}
For a fresh Gaussian draw independent of the actor output and $s<1$, the conditional law is
\begin{equation}
 z\mid(o,u)\sim\mathcal{N}\!\left(s\sqrt{N}\,v,(1-s^2)I_N\right).
\end{equation}
This uses the same relative weights on the mean and Gaussian noise as a variance-preserving diffusion step~\citep{ddpm}.
In the tiled $\pi_0$ configurations, retaining more Gaussian noise improves initial success (\cref{sec:initialization}).
The amount of Gaussian noise balances retention of base-policy behavior against the actor's control, motivating the policy-specific settings in \cref{tab:repro:steering}.
For tiled projections, mixing in Gaussian noise also weakens the repeated structure introduced across the action horizon.
This provides a possible explanation for their higher initial success at lower steering strength.
The random-projection initialization results add an important distinction: they retain near-base performance even at full steering strength ($s=1$; \cref{app:repro:action}), without fresh Gaussian noise.
Retaining a Gaussian component is therefore one way to preserve a useful starting point, while the structure of the actor-generated noise also matters.

\subsection{Action-Noise Steering}
\label{app:ablations}

\begin{figure}[tb]
  \centering
  \includegraphics{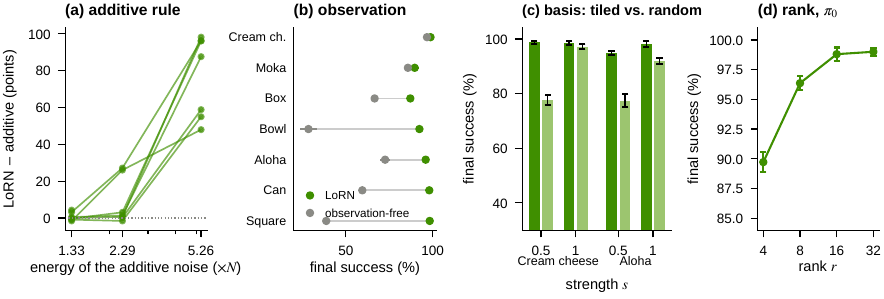}
\caption{\textbf{Noise scale, observation dependence, and projection structure affect action-noise steering.}
Ablations of \shortname with frozen $\pi_0$ on LIBERO and ALOHA and Diffusion Policy on Robomimic, five training seeds per configuration, with the final-success estimate of \cref{tab:action}.
Steering strength is the cap $\alpha$, which the actor's outputs saturate in these runs (\cref{app:repro:action}); full strength uses no fresh Gaussian component.
(a) \shortname success minus additive-steering success at matched strength, against the additive noise's expected squared norm divided by the Gaussian reference $N$; positive values favor \shortname.
(b) Success for an actor receiving observations (green) or zero inputs (gray), with critic observations unchanged.
(c) Tiled projections (dark green), which repeat one noise pattern across the action horizon, against random projections (light green) at matched rank: rank 16 at strength 0.5 and rank 32 at full strength on Cream cheese, rank 32 throughout on ALOHA Transfer Cube.
(d) Cream-cheese success against tiled projection rank at strength 0.5; gains level off beyond rank 16.}
  \label{fig:action_ablations}
\end{figure}

\paragraph{The gains appear in both learning progress and final success.}
\Cref{tab:action} gives the numbers behind the twelve learning curves of \cref{fig:action-policy-curves}, with the early-window and final-success summaries defined in \cref{app:repro:action}.
\Cref{fig:action-sample-efficiency} summarizes the same curves as the training steps each method needs to reach a sustained success target; \shortname reaches the target first on six of the twelve tasks, including all three Robomimic tasks.
\shortname has the highest early-window mean on nine tasks, including ties with LP-DS on Cream cheese and Pack container; DSRL leads on Alphabet soup, and LP-DS on Butter and Load dishwasher.
On $\pi_0$ and Diffusion Policy, final success is within three percentage points of DSRL on every task; on $\pi_{0.5}$, \shortname exceeds the stronger baseline by 19.9 percentage points on Load dishwasher, where it overtakes DSRL near 400k environment steps, and by 3.7 points on Pack container.

\begin{table}[tb]
    \centering
    \small
    \setlength{\tabcolsep}{4pt}
    \renewcommand{\arraystretch}{1.12}
\caption{\textbf{Action-noise steering improves learning progress and final success across robot policies.}
Success (\%) for frozen $\pi_0$ on six LIBERO tasks and ALOHA Transfer Cube, $\pi_{0.5}$ on two YAM tasks, and Diffusion Policy (DP) on Robomimic Lift, Can, and Square.
``Base'' is success without steering; DSRL replaces action noise, LP-DS adds a constrained offset, and \shortname steers noise through a normalized projection and Gaussian mixing.
Values are mean $\pm$ standard error over five training seeds.
Early-window means summarize success over 0--200k gradient updates for LIBERO, 0--1M for ALOHA, 0--500k environment steps for Lift and the YAM tasks, and 0--1M for Can and Square: time-normalized areas under the success curves for $\pi_0$ (five-evaluation running means) and DP (raw evaluations), and the average of the evenly spaced raw evaluations for $\pi_{0.5}$ (\cref{app:repro:action}).
Final success averages each seed's last ten evaluations, taken on the five-evaluation running mean for $\pi_0$; higher values are better in both groups.
Bold marks the highest mean among the three methods in each group; ties are both bold.
\shortname has the highest early-window mean on nine tasks, including ties, and the highest final mean on both $\pi_{0.5}$ tasks.
LP-DS uses the released implementation for DP and policy-specific adaptations for $\pi_0$ and $\pi_{0.5}$.}
    \label{tab:action}
    \setlength{\tabcolsep}{3pt}
    \renewcommand{\arraystretch}{1.12}
    \begin{tabular}{llc ccc ccc}
        \toprule
        & & & \multicolumn{3}{c}{Early-window mean (\%) $\uparrow$} & \multicolumn{3}{c}{Final success (\%) $\uparrow$} \\
        \cmidrule(lr){4-6}\cmidrule(lr){7-9}
        Policy & Task & Base & DSRL & LP-DS & \shortname & DSRL & LP-DS & \shortname \\
        \midrule
        \multirow{7}{*}{$\pi_0$}
          & Cream cheese & 29.3 & 81.1 $\pm$ 2.0 & \textbf{88.2 $\pm$ 1.8} & \textbf{88.2 $\pm$ 1.0} & \textbf{98.8 $\pm$ 0.8} & 97.0 $\pm$ 1.1 & \textbf{98.8 $\pm$ 0.6} \\
          & Moka pot & 80.7 & 63.6 $\pm$ 2.7 & 66.7 $\pm$ 3.1 & \textbf{70.1 $\pm$ 2.4} & \textbf{91.1 $\pm$ 3.1} & 78.0 $\pm$ 5.1 & 89.7 $\pm$ 2.4 \\
          & Cream-cheese box & 60.7 & 55.3 $\pm$ 5.1 & 51.6 $\pm$ 4.6 & \textbf{66.1 $\pm$ 1.6} & 84.3 $\pm$ 2.0 & 64.7 $\pm$ 5.4 & \textbf{87.2 $\pm$ 1.9} \\
          & Black bowl & 0.4 & 44.7 $\pm$ 4.8 & 28.7 $\pm$ 5.1 & \textbf{53.0 $\pm$ 4.3} & \textbf{93.8 $\pm$ 3.0} & 64.5 $\pm$ 5.6 & 92.4 $\pm$ 1.5 \\
          & Alphabet soup & 51.1 & \textbf{89.2 $\pm$ 0.9} & 86.6 $\pm$ 1.8 & 79.9 $\pm$ 2.9 & \textbf{98.4 $\pm$ 0.5} & 94.2 $\pm$ 1.1 & 96.4 $\pm$ 0.6 \\
          & Butter & 10.0 & 58.5 $\pm$ 2.3 & \textbf{65.2 $\pm$ 6.4} & 60.8 $\pm$ 5.1 & 91.6 $\pm$ 0.9 & 87.0 $\pm$ 3.4 & \textbf{93.4 $\pm$ 1.0} \\
          & Transfer Cube & 71.2 & 82.4 $\pm$ 0.7 & 85.4 $\pm$ 1.4 & \textbf{88.2 $\pm$ 0.4} & \textbf{98.4 $\pm$ 0.4} & 94.1 $\pm$ 0.7 & 96.0 $\pm$ 1.1 \\
        \midrule
        \multirow{2}{*}{$\pi_{0.5}$}
          & Load dishwasher & 18.8 & 15.0 $\pm$ 1.3 & \textbf{28.3 $\pm$ 0.9} & 12.5 $\pm$ 1.1 & 49.6 $\pm$ 2.5 & 39.8 $\pm$ 1.9 & \textbf{69.5 $\pm$ 2.9} \\
          & Pack container & 13.3 & 4.5 $\pm$ 0.7 & \textbf{13.8 $\pm$ 1.6} & \textbf{13.8 $\pm$ 0.8} & 2.8 $\pm$ 0.8 & 18.9 $\pm$ 2.0 & \textbf{22.6 $\pm$ 1.5} \\
        \midrule
        \multirow{3}{*}{DP}
          & Lift & 81.0 & 86.4 $\pm$ 1.2 & 94.4 $\pm$ 0.2 & \textbf{94.7 $\pm$ 0.6} & 99.6 $\pm$ 0.1 & 99.2 $\pm$ 0.1 & \textbf{99.8 $\pm$ 0.1} \\
          & Can & 39.0 & 79.4 $\pm$ 1.2 & 80.7 $\pm$ 0.8 & \textbf{86.8 $\pm$ 0.6} & 96.0 $\pm$ 0.4 & 98.0 $\pm$ 0.2 & \textbf{98.1 $\pm$ 0.2} \\
          & Square & 50.5 & 59.6 $\pm$ 7.8 & 69.2 $\pm$ 0.9 & \textbf{79.6 $\pm$ 0.6} & 97.3 $\pm$ 0.5 & \textbf{98.3 $\pm$ 0.2} & \textbf{98.3 $\pm$ 0.1} \\
        \bottomrule
    \end{tabular}
\end{table}

The following controls examine additive steering, observation dependence, projection structure, and rank (\cref{fig:action_ablations}).
Each configuration uses five training seeds and the final-success estimate in \cref{tab:action}.

\paragraph{Larger norm changes can erase the benefit of steering.}
At maximum strengths $\alpha$ of 0.5, 0.75, and 0.9, which the actor's outputs saturate throughout these runs so that the delivered strength equals $\alpha$, the additive construction of \cref{eq:additive_control} has expected squared norm approximately 1.33, 2.29, and 5.26 times the Gaussian reference (root-mean-square norm 1.15, 1.51, and 2.29 times $\sqrt{N}$).
At the smallest increase it matches \shortname on every task; at the intermediate increase it still matches on five of seven tasks but trails by 26 and 27 points on Moka pot and Cream-cheese box; at the largest it falls below the base policy on every task.
Norm preservation therefore matters most when substantial steering is needed, and the task differences at intermediate strength agree with the amplitude probes.

\paragraph{The observation gives the actor useful information for steering.}
Zeroing the actor's observations lowers final success by 2--4 percentage points on Cream cheese and Moka pot and by 21--64 points on the other five tasks: a single favorable noise distribution is not enough, and the size of the gap varies by task.
\Cref{sec:actor_directions} shows how this observation dependence develops during training.

\paragraph{Projection structure matters even when rank is matched.}
At steering strength 0.5, tiled projections beat random projections of the same rank by 21 percentage points on Cream cheese and 18 on ALOHA Transfer Cube; at full strength the gaps narrow to one and six points.
Tiling imposes a shared pattern across the horizon, which appears to help learning when combined with Gaussian noise, whereas random projections preserve the initial success better at full strength (\cref{sec:initialization}): a good starting structure and a good learning structure need not coincide.

\paragraph{A compact tiled projection is enough on Cream cheese.}
Tiled ranks 4, 8, 16, and 32 reach 89.7\%, 96.4\%, 98.8\%, and 99.0\%, which motivates rank 16 for $\pi_0$; this is the action-policy counterpart of the Cosmos3-Edge rank sweep, not a claim that one rank suits every policy.

\subsection{Observation-Dependent Steering}
\label{sec:actor_directions}

\paragraph{Analysis protocol.}
We examine how the actor's use of its subspace develops during Cosmos3-Edge video training on Mustard and Marker.
For each checkpoint, we evaluate the deterministic normalized output $\hat{u}=u/\|u\|$, with $u=\tanh(\mu(o))$, on the observations recorded in one 240-episode evaluation of the final policy: 2,286 steering decisions on Mustard at 1.5M environment steps and 3,814 on Marker at 966k.
The actor receives eight joint/gripper coordinates and an encoding of the real conditioning frame from the previous decision.
We measure how much the output depends on the observation as the spread $1-c^2$, where $c=\|n^{-1}\sum_i\hat{u}_i\|$ over the $n$ recorded observations; zero spread means the actor selects the same normalized output for every observation.

\begin{figure}[H]
  \centering
  \includegraphics{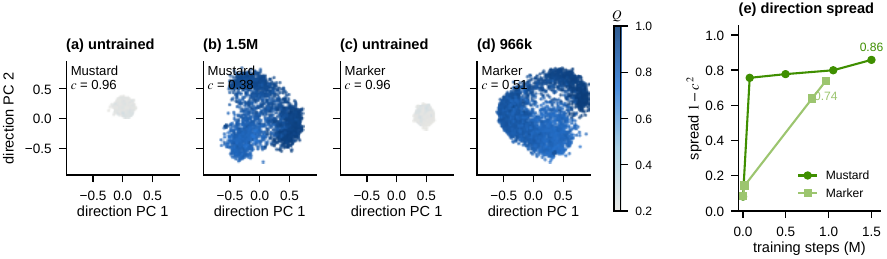}
\caption{\textbf{Training turns a nearly fixed noise choice into an observation-dependent strategy.}
Cosmos3-Edge rank-32 \shortname video steering on RoboLab MustardInLeftBin and MarkerInMug.
(a--d) Normalized actor outputs $\hat{u}$ of the untrained and final actors, evaluated deterministically on the same observations from a 240-episode evaluation of the final policy and projected onto the first two principal components of the final actor's centered outputs, fitted separately per task.
Point color gives the mean value of ten critics; $c$ is the norm of the mean output.
(e) Spread $1-c^2$ over training, computed before PCA; zero spread means identical normalized outputs for every observation.
Final checkpoints are 1.5M environment steps for Mustard and 966k for Marker.}
  \label{fig:cloud_training}
\end{figure}

\paragraph{Training turns a nearly fixed noise choice into an observation-dependent one.}
At initialization the spread is 0.08 on Mustard: the untrained actor selects nearly the same noise in every situation.
It reaches 0.76 by 77k environment steps and 0.86 at 1.5M; on Marker it rises later, reaching 0.64 at 806k steps and 0.74 at 966k (\cref{fig:cloud_training}).
Because the observations are fixed and the outputs are deterministic, these changes reflect the actor's learned response to observations rather than new visited states or exploration noise.

Together with the zero-observation control of \cref{fig:action_ablations}, the result supports an adaptive use of the fixed subspace rather than the selection of one noise input for every situation.
The spread measures how much the noise choice depends on the observation; whether the learned choices improve behavior is established by the success evaluations, not by the spread.

\subsection{YAM Station Trials}
\label{app:real_trials}

\begin{figure}[p]
  \centering
  \includegraphics{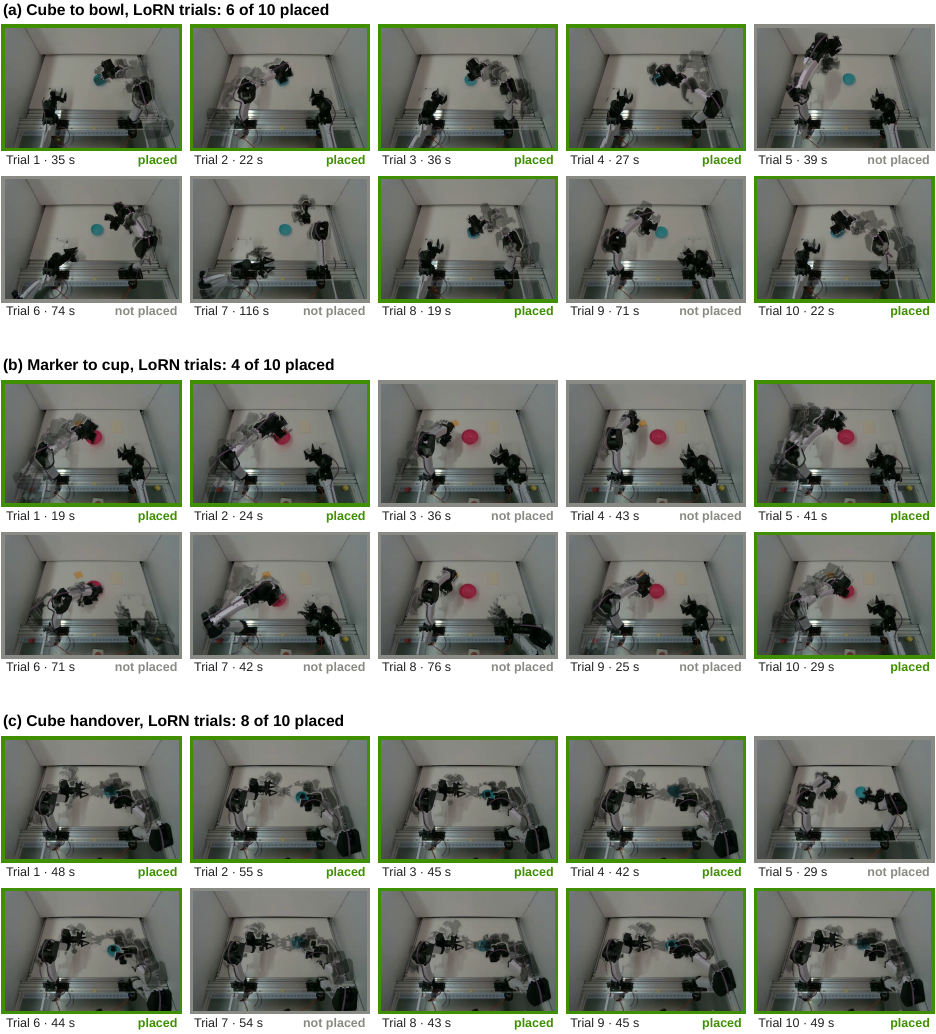}
\caption{\textbf{Simulation-trained steering transfers to physical placement tasks on the YAM station.}
All ten trials per task of a simulation-selected \shortname actor steering the action noise of the frozen policy, with no actor or policy training on hardware: GR00T N1.7 in (a) Cube to bowl and (b) Marker to cup, and the fine-tuned $\pi_{0.5}$ in (c) Cube handover, shown in recording order.
Each panel uses the initial overhead frame as its background, overlays six intermediate arm poses translucently, and shows the final pose solid; labels give trial durations.
Green borders mark successful placement, where the released object remains inside the target container through the final second; gray borders mark no placement.
\shortname succeeds in 6 of 10 cube trials, 4 of 10 marker trials, and 8 of 10 handover trials.}
  \label{fig:real_yam}
\end{figure}

\paragraph{Simulation-trained steering transfers directly to the YAM station.}
\Cref{fig:yam_tasks} shows the three tasks from the overhead camera.
\Cref{fig:real_yam} presents all \shortname physical trials reported in \cref{tab:real_yam}, in recording order.
The evaluation protocol and outcome definitions appear in \cref{app:repro:real}.
The successful physical trials show that an actor trained against the simulated policy can select noise that remains useful on the YAM station.
Because the base-policy weights remain frozen and the actor receives no hardware training, this transfer operates through the learned noise input.
It suggests that successful behavior is already accessible to the frozen model, while the steering actor learns how to elicit it from observations.
The complete trial sequence shows the observed successes and unsuccessful attempts, providing context for the aggregate placement counts.

\clearpage

\end{document}